\documentclass[11pt,a4paper]{article}
\usepackage[hyperref]{emnlp-ijcnlp-2019}
\usepackage{times}
\usepackage{latexsym}
\usepackage{amsmath}
\usepackage{amssymb}
\usepackage{dsfont}

\usepackage{booktabs}
\usepackage{graphicx}
\usepackage{url}

\aclfinalcopy 

\title{Emergent Linguistic Phenomena in Multi-Agent Communication Games}

\author{Laura Graesser$^{\dagger *}$, Kyunghyun Cho$^{\dagger \ddagger \star}$, Douwe Kiela$^\ddagger$\\
$^\dagger$ NYU; $^*$ Robotics at Google; $^\star$ CIFAR Azrieli Global Scholar; $^\ddagger$ Facebook AI Research\\
lauragraesser@google.com, kyunghyun.cho@nyu.edu, dkiela@fb.com
}

\date{}

\begin{document}
\maketitle
\begin{abstract}
We describe a multi-agent communication framework for examining high-level linguistic phenomena at the community-level. We demonstrate that complex linguistic behavior observed in natural language can be reproduced in this simple setting: i) the outcome of contact between communities is a function of inter- and intra-group connectivity; ii) linguistic contact either converges to the majority protocol, or in balanced cases leads to novel creole languages of lower complexity; and iii) a linguistic continuum emerges where neighboring languages are more mutually intelligible than farther removed languages. We conclude that at least some of the intricate properties of language evolution need not depend on complex evolved linguistic capabilities, but can emerge from simple social exchanges between perceptually-enabled agents playing communication games.
\end{abstract}

\section{Introduction}

Contact linguistics \cite{myers2002contact} studies what happens when two or more languages or language varieties interact. It poses several pertinent open questions that are difficult to answer: how does symmetric (``mutually intelligible'') communication emerge; how do languages behave under contact; how does one language come to dominate another; how and why does extensive language contact tend to lead to simplification (e.g. in creoles); and how does a linguistic continuum come about, where neighboring languages are more intelligible than farther removed ones? In this work, we show that such linguistic phenomena emerge naturally given a few general assumptions about the organizational structure of networks of artificial agents equipped with a minimalistic form of learned communication.

We introduce a multi-agent framework for studying the emergence and evolution of language, where agents are neural networks endowed with the ability to exchange messages about their perceptual input. The advantage of this approach is that one can precisely control linguistic, environmental and algorithmic variables.

First, we investigate linguistic behavior at the agent-level, and examine when symmetric communication emerges within a linguistic community, as well as how the topological organization of communities---i.e., which other agents an agent comes into contact with and how frequently that happens---impacts convergence and learning. We then examine the behavior of communities of such agents when they come into contact, as well as how community-level topology impacts convergence, success rate and mutual intelligibility.

We demonstrate that the following linguistic behaviors emerge, which correspond to known linguistic phenomena in natural languages: 1) the outcome of contact is a function of inter- and intra-group connectivity, i.e. that languages become mutually intelligible through contact, even for agents that have not themselves been exposed to the other language, provided there is sufficient connectivity between communities; 2) linguistic contact over time either converges to the dominant majority protocol, leading to the extinction of the other language, or if the communities are balanced, gives rise to an original ``creole'' protocol that has lower complexity than the original languages; 3) a linguistic continuum emerges, where neighboring languages are more mutually intelligible than farther removed languages and the topology of the continuum governs its behavior. 

To our knowledge, this work constitutes the first attempt at studying contact linguistic phenomena using communities of deep neural agents. Our findings indicate that intricate properties of language evolution need not depend on intrinsic properties of highly complex evolved linguistic capabilities, but instead can emerge purely from social exchanges between perceptually-enabled agents with simple communicative capabilities.

\section{Related Work}

Studying language change {\it in vivo} is challenging, since it requires simultaneous observation of speaker and community interactions~\cite{brooks2008prolonged,trudgill1974linguistic,joseph2017,christiansen2003language}, while carefully controlling for purposes and goals \cite{winograd1971procedures,winogradflores87,nowak1999evolution}. Studies of language emergence and evolution must furthermore be conducted over a long period of time, spanning decades or even centuries. Even the Nicaraguan sign language, which emerged remarkably rapidly, took several decades to develop fully~\cite{senghas2005emergence}.
Language itself also never ceases to evolve~\cite{fishman1964language}.

Advances in computer science have provided us with opportunities for instead investigating the emergence and evolution of languages {\it in vitro} using computational and mathematical models~\cite{hurford:1989saussurean,briscoe2002linguistic,kirby2002natural,christiansen2003language,kirby2008cumulative,lewis2008convention,skyrms2010signals}. In computational approaches, communities of agents, equipped with the ability to communicate, are deployed in a simulated environment. Their communication protocol is either evolved or learned, in order to maximize some reward provided by the environment. The agents' behavior and communication are observed and used to compare against linguistic phenomena or hypothesized linguistic theories.

Computational multi-agent models are characterized by the complexity of the agents, the choice of learning algorithm, and the design of the environment and reward structure. The complexity of an artificial agent ranges from a set of simple difference equations~\cite{grouchy2016evolutionary}, to a CPU-like architecture with an instruction set and registers~\cite{knoester2007directed}, to a co-occurrence matrix between objects and symbols~\cite{nowak1999evolution}, to a simple single-layer neural network~\cite{trianni2006self}, to a deep neural network~\cite{lazaridou2016multi,foerster2016learning,jorge2016learning}. The learning algorithm is either a variant of evolutionary algorithms~\cite{nowak1999evolution,kirby2002natural,grouchy2016evolutionary}, often used in the framework of Artificial Life~\cite{bedau2003artificial}, or a gradient-based optimization algorithm, as is often used for training deep neural networks with a supervised or reinforcement learning objective function. The former simulates generations developing complex behavior over time, while the latter enables more sophisticated agents thanks to the recent advances in deep learning~\cite{lecun2015deep}. Recent years have seen intriguing new results in emergent communication, starting with~\newcite{lazaridou2016multi} and~\newcite{foerster2016learning}, using 
deep neural agents~\cite{lewis2017deal,havrylov2017emergence,jorge2016learning,evtimova2018emergent,Das_2018_CVPR,cao2018emergent}. Often, these approaches could be framed as special or generalized cases of Lewis's signalling game~\cite{lewis2008convention}, in which agents exchange signals to achieve a common goal. In this work, deep neural agents play games within communities of similar agents, where the aim is for agents to communicate about their perceptual input.

\begin{figure*}[t]
\centering
\raisebox{-0.5\height}{\includegraphics[width=.3\textwidth]{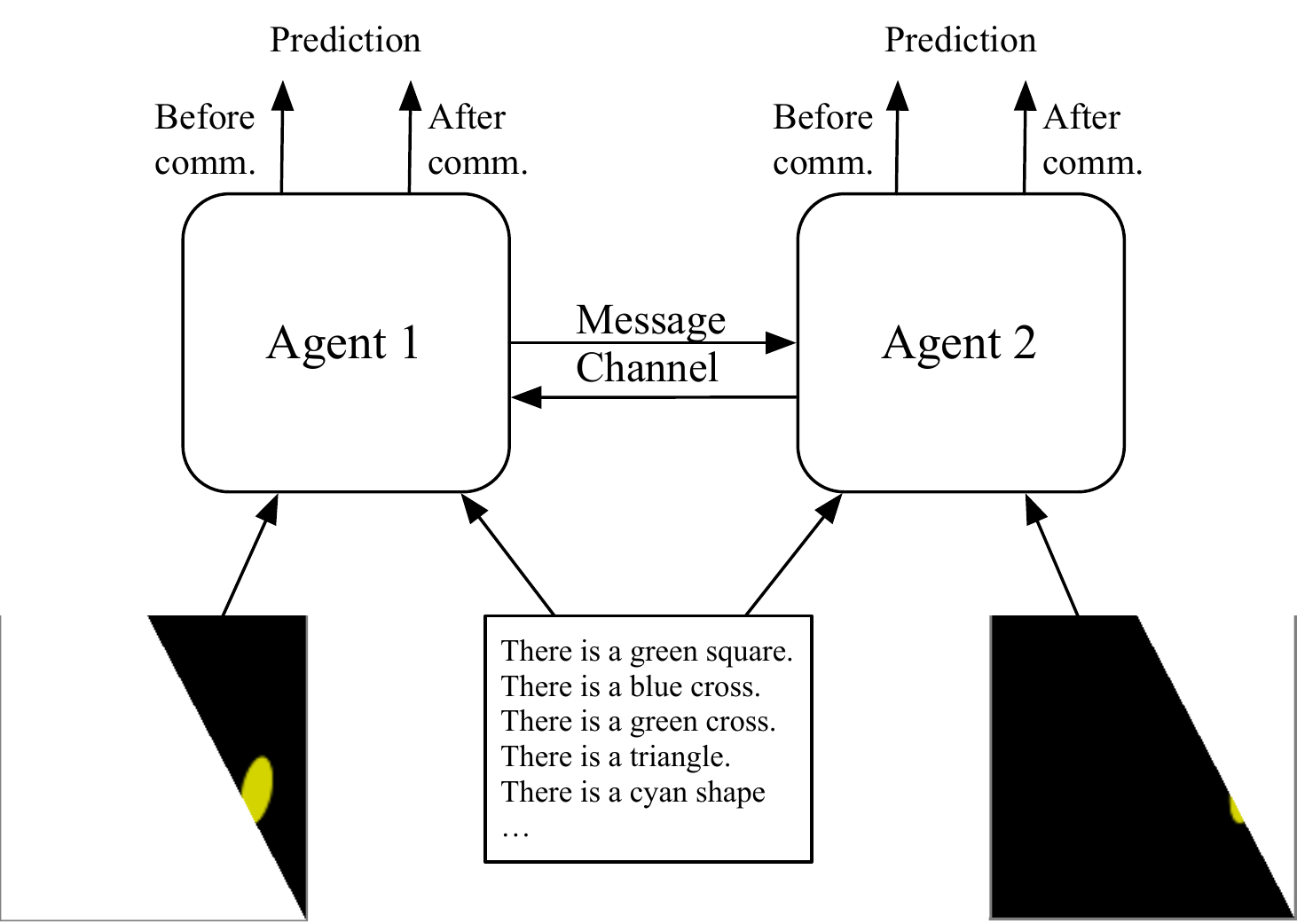}}\hfill
\raisebox{-0.5\height}{\includegraphics[width=.3\textwidth]{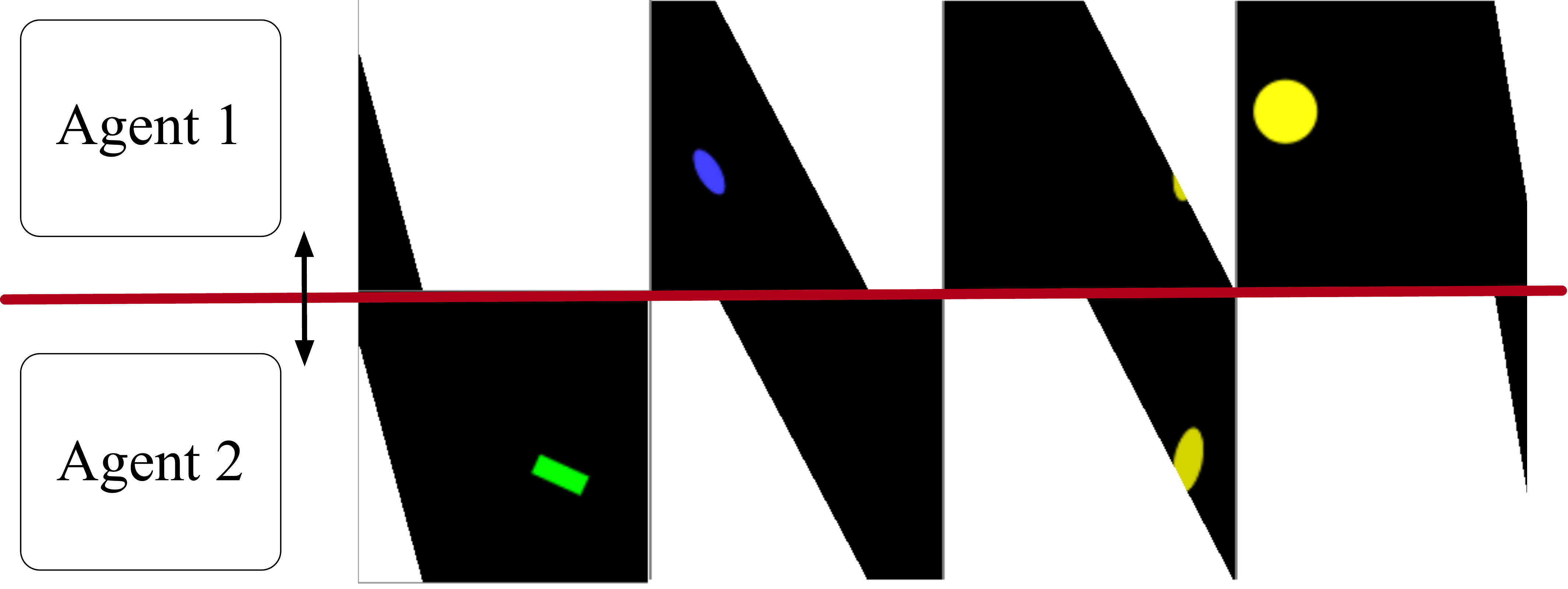}}\hfill
\raisebox{-0.5\height}{\includegraphics[width=.3\textwidth]{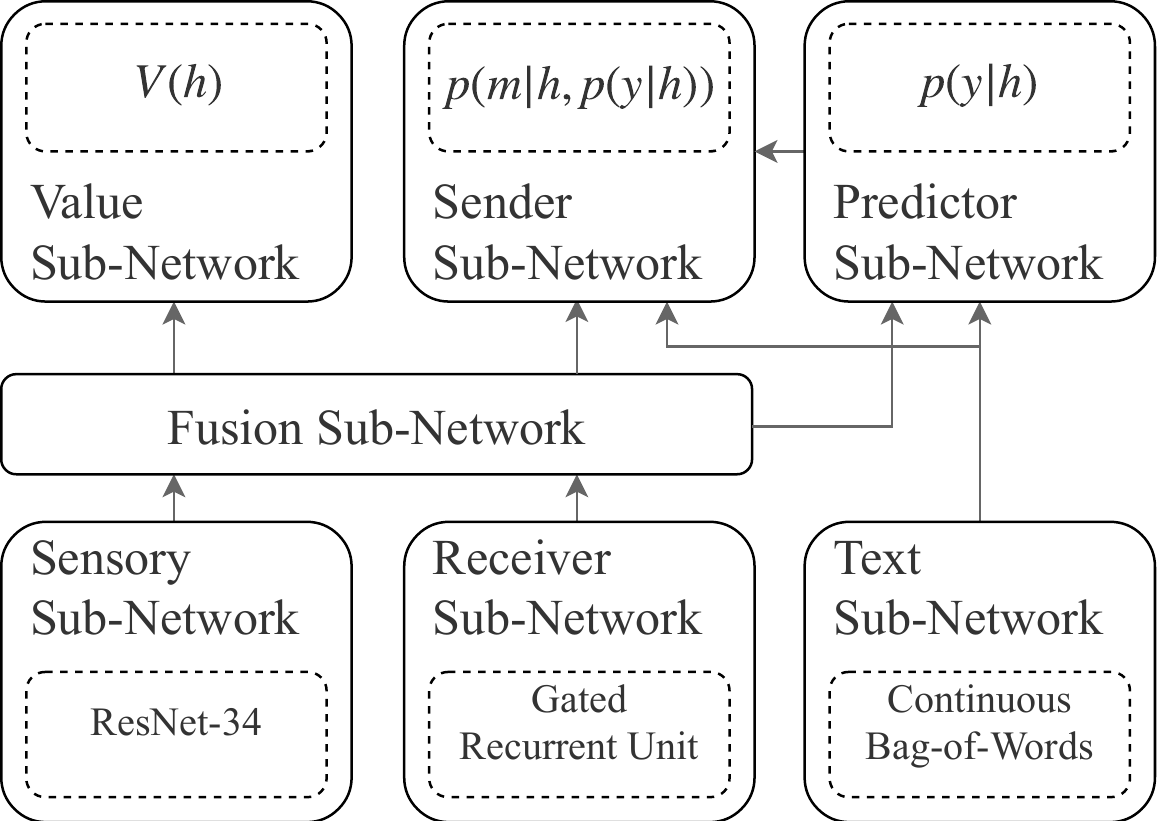}}

\caption{
\label{fig:agent}
Overview of the communication game and agent structure.
\textbf{Left}: A graphical illustration of the proposed game. Each of two agents observes a partition of an input image and decides which of ten textual captions best describes the entire image before and after exchanging messages with the other agent. \textbf{Middle}: Example training data. Only a random part of each image (dark background) is presented to one agent, necessitating communication in order to solve the game.  \textbf{Right}: The modular structure of an agent.}
\end{figure*}

\begin{figure}
    \centering
    \begin{minipage}{\columnwidth}
    \centering
    \includegraphics[width=.48\columnwidth]{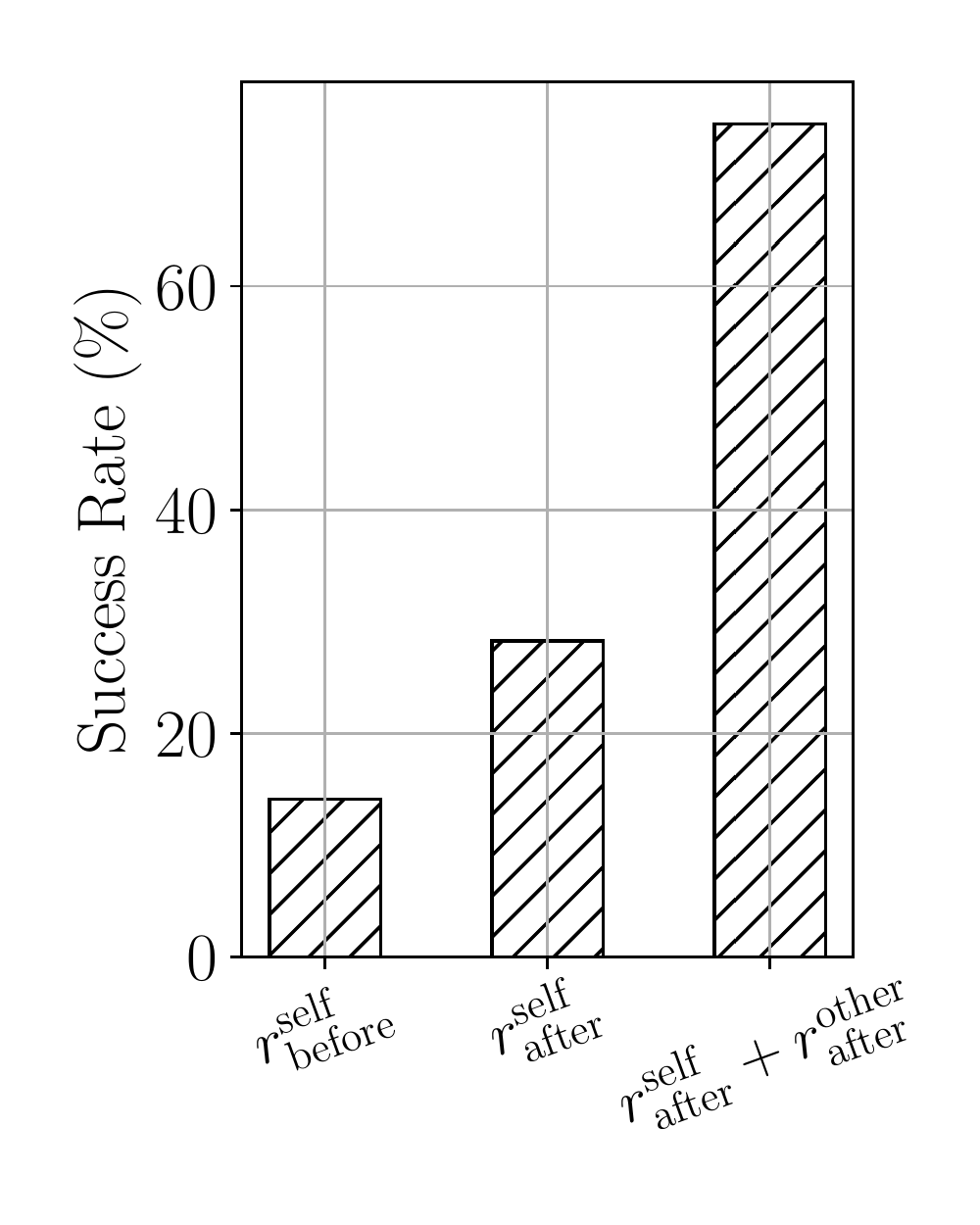}
    \includegraphics[width=.48\columnwidth]{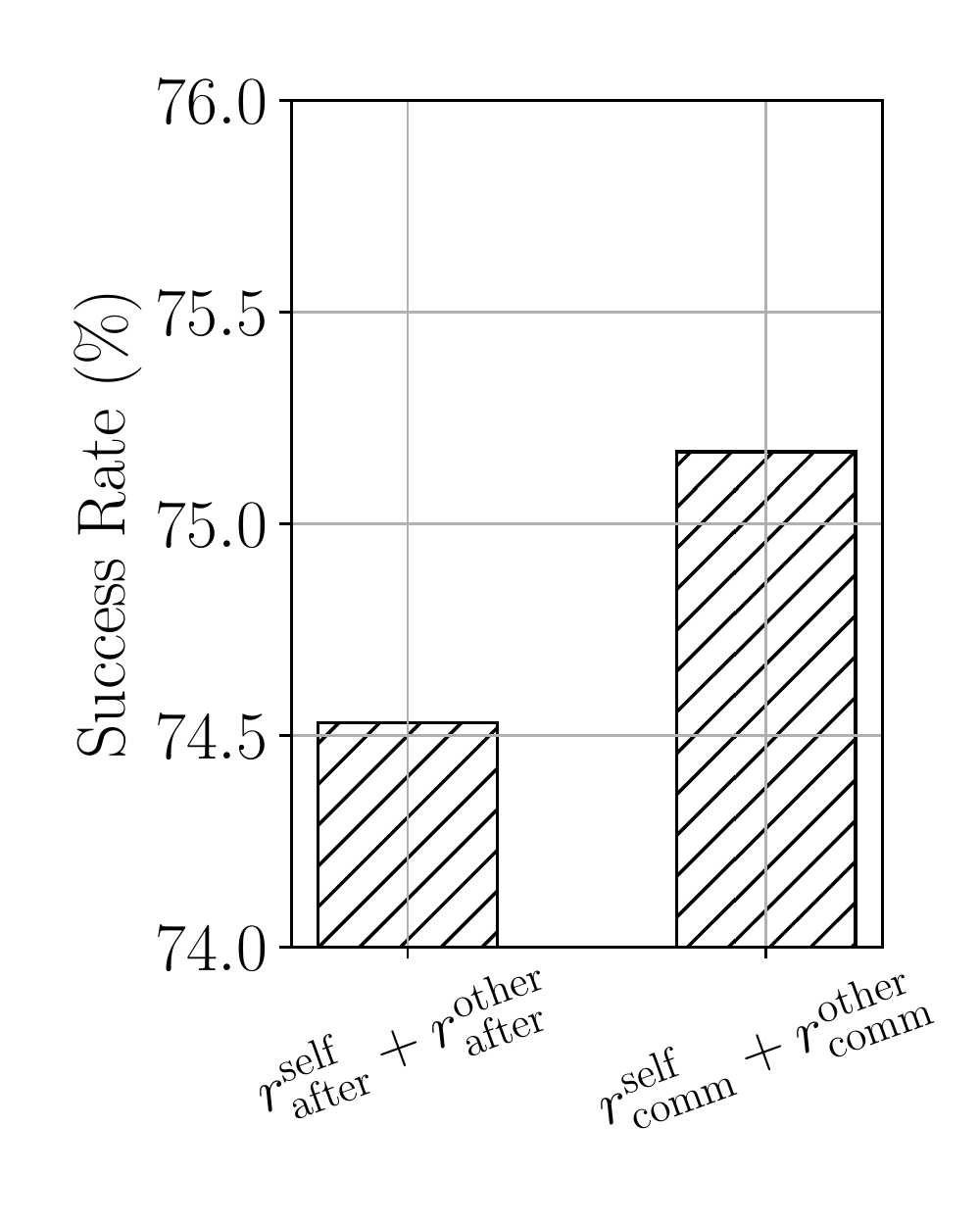}
    \end{minipage}
    \caption{\label{fig:reward}
    The chance of both agents correctly guessing the answer depends on the reward (averaged  over  three  runs): (\textbf{left}) it is important to reward the collective behaviour in order for two agents to collaborate; (\textbf{right}) the success rate goes up when we encourage agents to maximize the accuracy \emph{improvement} after communication.}
\end{figure}

\section{Multi-agent communication}

In order to study emergent linguistic phenomena in a simplified but realistic setting, the communication game needs to have several properties. First, it should be \emph{symmetric}, in that all agents should be able to act as ``speaker'' and ``listener''. Second, the agents should communicate about something \emph{external} to themselves, i.e., about the sensory experience of something in their environment. Third, the world should be \emph{partially observable}, implying that communication is required for solving the game successfully. {\bf Fig.~\ref{fig:agent}} shows example training data, the game setup and agent design. See the supplementary material for details.

\paragraph{Reference Game}

Let $G=\langle A, O, M, W, R\rangle$ be a multi-agent communication game, consisting of communities of agents~$A$ communicating across the bidirectional message channel~$M$ given environmental observations~$O$, where the community membership of agents is defined as a graph with the agents $A$ as vertices and weighted edges~$W$ that determine whether two agents are connected, i.e.,~$W$ specifies the topology of the network. 

Each agent is a deep neural network with perceptual inputs and given a language description, as illustrated in {\bf Fig.\ref{fig:agent}}. All agents have the same network architecture because of the requirement that the game be symmetric, but each agent has its own parameters which are learned during training. Pairs of connected agents learn to play a game with a reward structure $R$, designed specifically to require communication-based collaboration. The exchange of information through the communication channel~$M$ can take any form. By learning to play the game, agents develop a communication protocol about the observations. This framework allows us to control for proximity constraints, population size and degree of interaction, through the underlying graph structure.

Throughout the game, each agent observes one part of an image that contains an object of a specific shape and color and is given a set of $n_{\text{cap}}$ synthetic compositional captions (e.g. ``there is a green triangle'') of which only one correctly describes the image\footnote{ShapeWorld~\cite{kuhnle2017shapeworld} was used to derive the data set which includes both the images and captions.
See the supplementary material for details.
}.

The goal is for the agents to identify the correct caption $y^*$ for the image. Since each agent only has partial information, the pair of agents must cooperate via communication to be effective at solving the problem {\it together}. 

At the beginning of the game, each agent makes an initial guess $\hat{y}_0$ of the correct answer, followed by $k$ rounds of communication in which the agents take turns transmitting a binary message to the other agent.
In the experiments discussed in this paper, agents communicate using 8-bit binary message vectors. Binary message vectors have been used before for studying the emergence and evolution of language~\cite{Kirby2002}. While we selected this type of communication for efficiency reasons, it is straightforward to replace it with sequences of discrete symbols~\cite{jorge2016learning,havrylov2017emergence,lee2017emergent,cao2018emergent,lazaridou2018emergence}, continuous vectors or larger binary vectors. After several rounds of communication, each agent makes their final guess $\hat{y}_1$. The game is considered successful if {\it both} of the agents correctly guessed the answer.

During training, a pair of agents corresponding to adjacent vertices, $a_i \in A$ and $a_j \in A$, is selected at random according to interaction weights $w_{ij} \in W$. 
One of them is randomly picked as the starting agent. 
The agents then play one instance of the reference game and their parameters are updated accordingly. The community structure can change during training. For instance, we are able to merge separately trained linguistic communities into a single community and fine-tune the agents from both communities (i.e. bring the communities into contact) to investigate linguistic contact. Once training is done, we can test pairs of distant agents to understand what changed in the protocol.

\paragraph{Reward}

The reward structure for each agent is designed as follows. First, we reward the agent when it correctly guesses the answer {\it after} communication $r^{\text{self}}_{\text{after}} = \mathds{1}_{y^* = \hat{y}_1}$, where $\mathds{1}$ is an indicator function, 
in order to encourage it to incorporate information received from the other agent. Second, we reward cooperation by giving each agent a shared reward composed of both its own and the other agent's rewards, i.e., $r_{\text{after}} = r^{\text{self}}_{\text{after}} + r^{\text{other}}_{\text{after}}$. As Fig.~\ref{fig:reward} shows, we empirically validate the importance of rewarding after-communication behavior and observe that the cooperation reward significantly boosts the success rate. Lastly, we explicitly encourage the agents to rely on communication by rewarding them for {\it relative improvement from communication}, rather than the success after communication: $r = r^{\text{self}}_{\text{comm}} + r^{\text{other}}_{\text{comm}}$, where 
$r^{\text{self}}_{\text{comm}} = r_{\text{after}}^{\text{self}} - 
r_{\text{before}}^{\text{self}}$ 
and 
$r^{\text{other}}_{\text{comm}} = r_{\text{after}}^{\text{other}} - 
r_{\text{before}}^{\text{other}}$. This final reward, which encourages both cooperation and explicit reliance on communication, reaches the highest success rate.

\begin{figure*}[t]
\centering
\begin{minipage}{0.35\textwidth}
\centering
\includegraphics[width=\columnwidth]{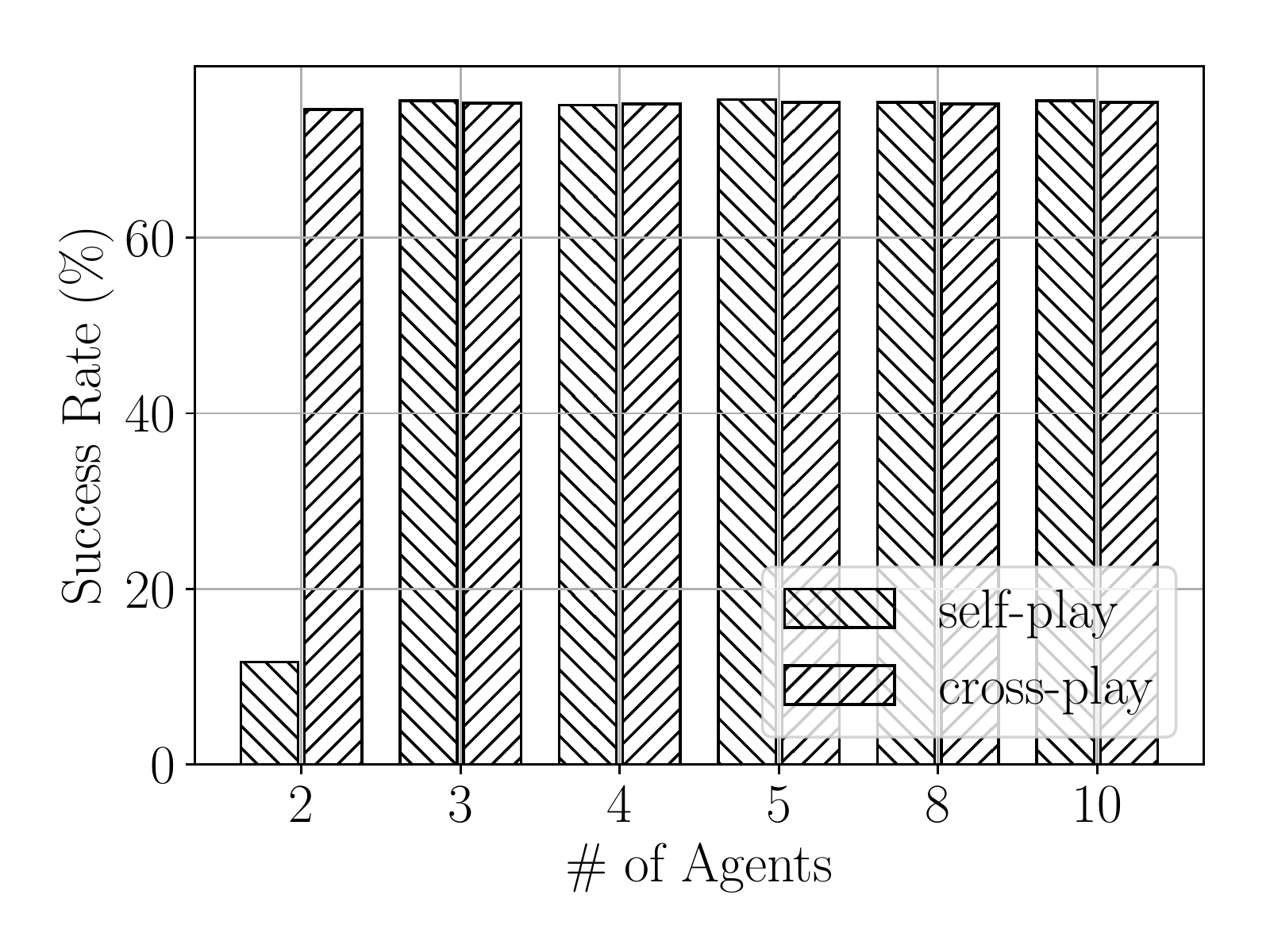}
(a)
\end{minipage}
\begin{minipage}{0.35\textwidth}
\centering
\includegraphics[width=\columnwidth]{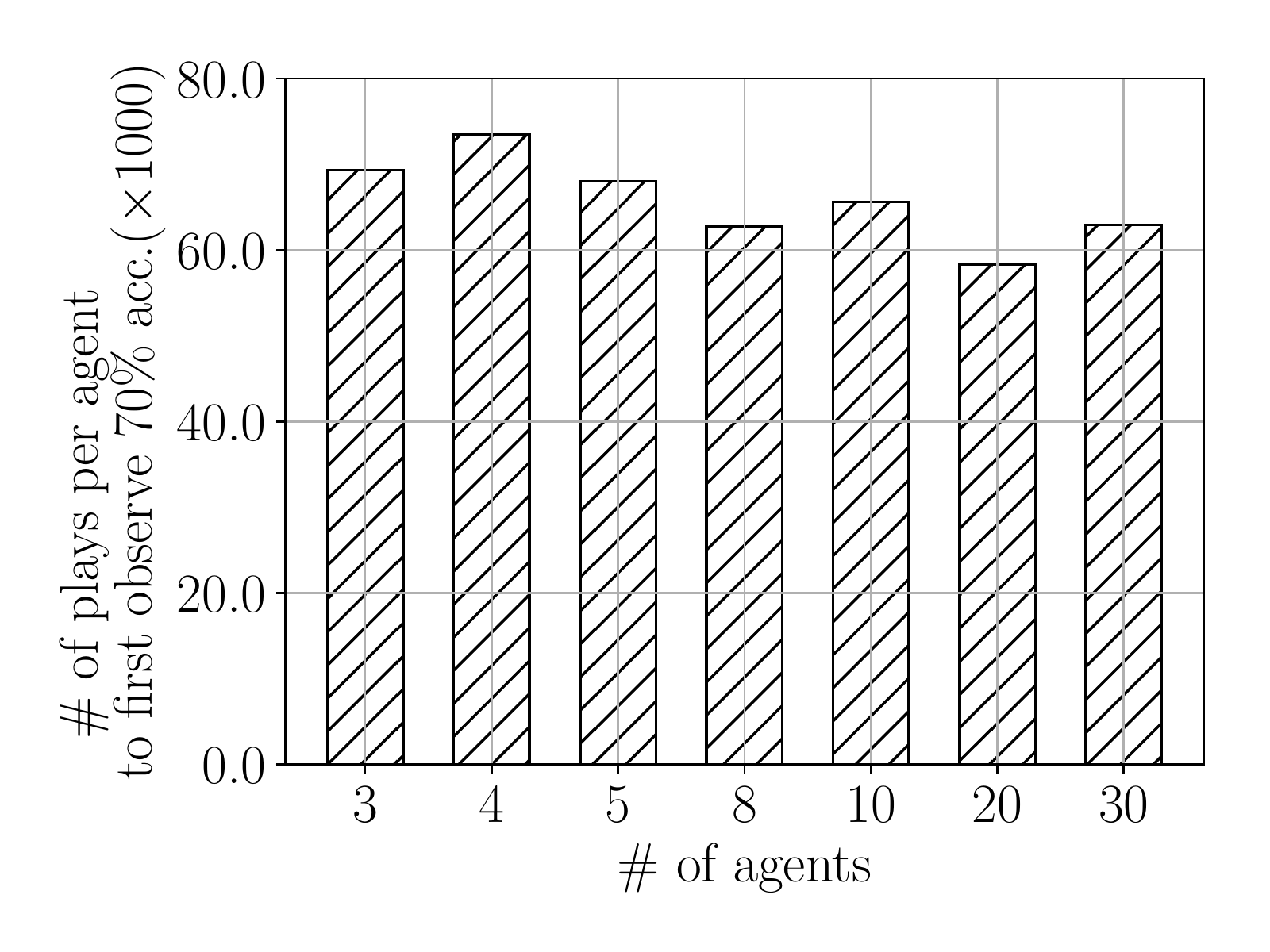}
(b)
\end{minipage}
\begin{minipage}{0.35\textwidth}
\centering
\includegraphics[width=\columnwidth]{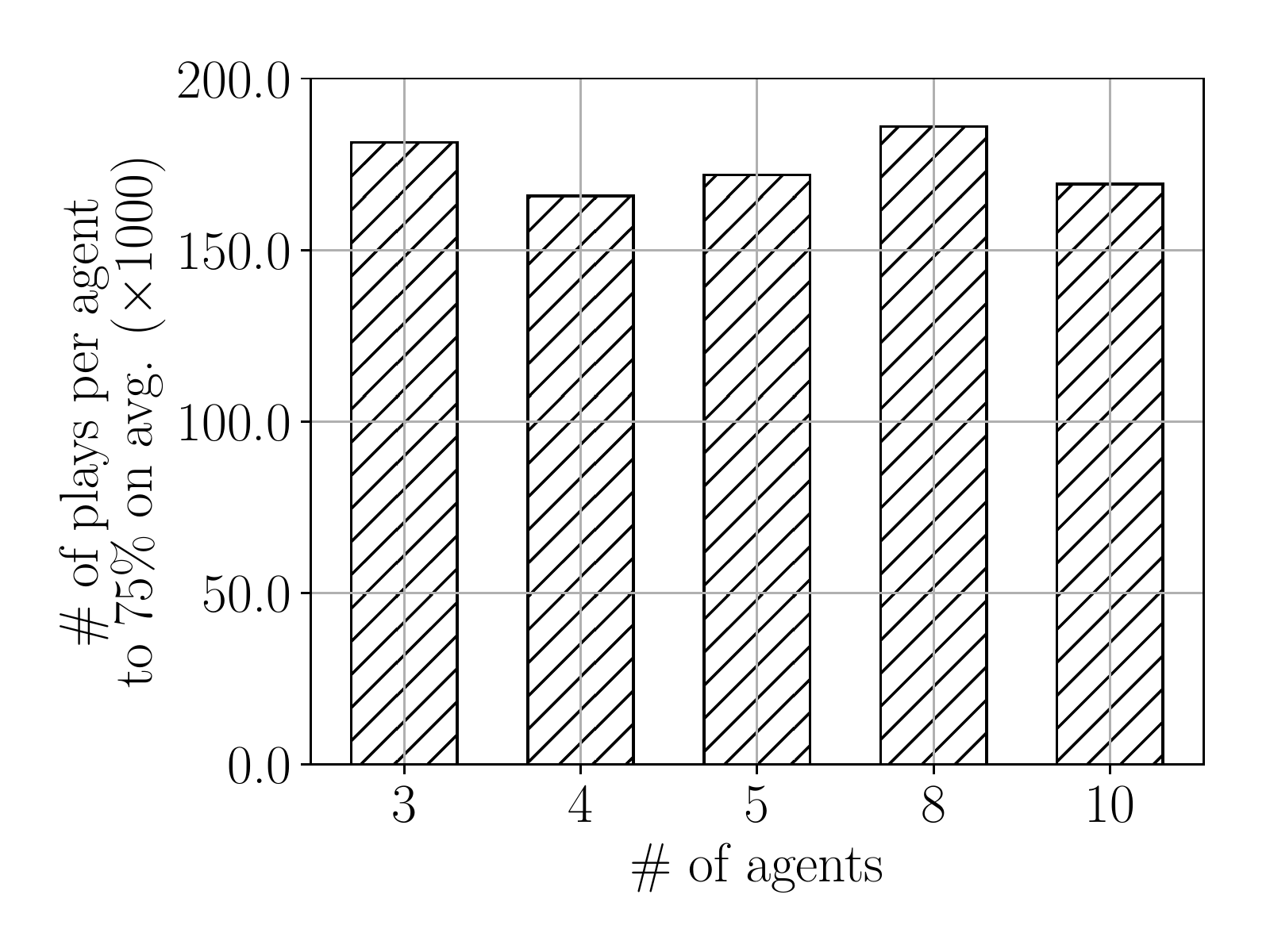}
(c)
\end{minipage}
\begin{minipage}{0.35\textwidth}
\centering
\includegraphics[width=\columnwidth]{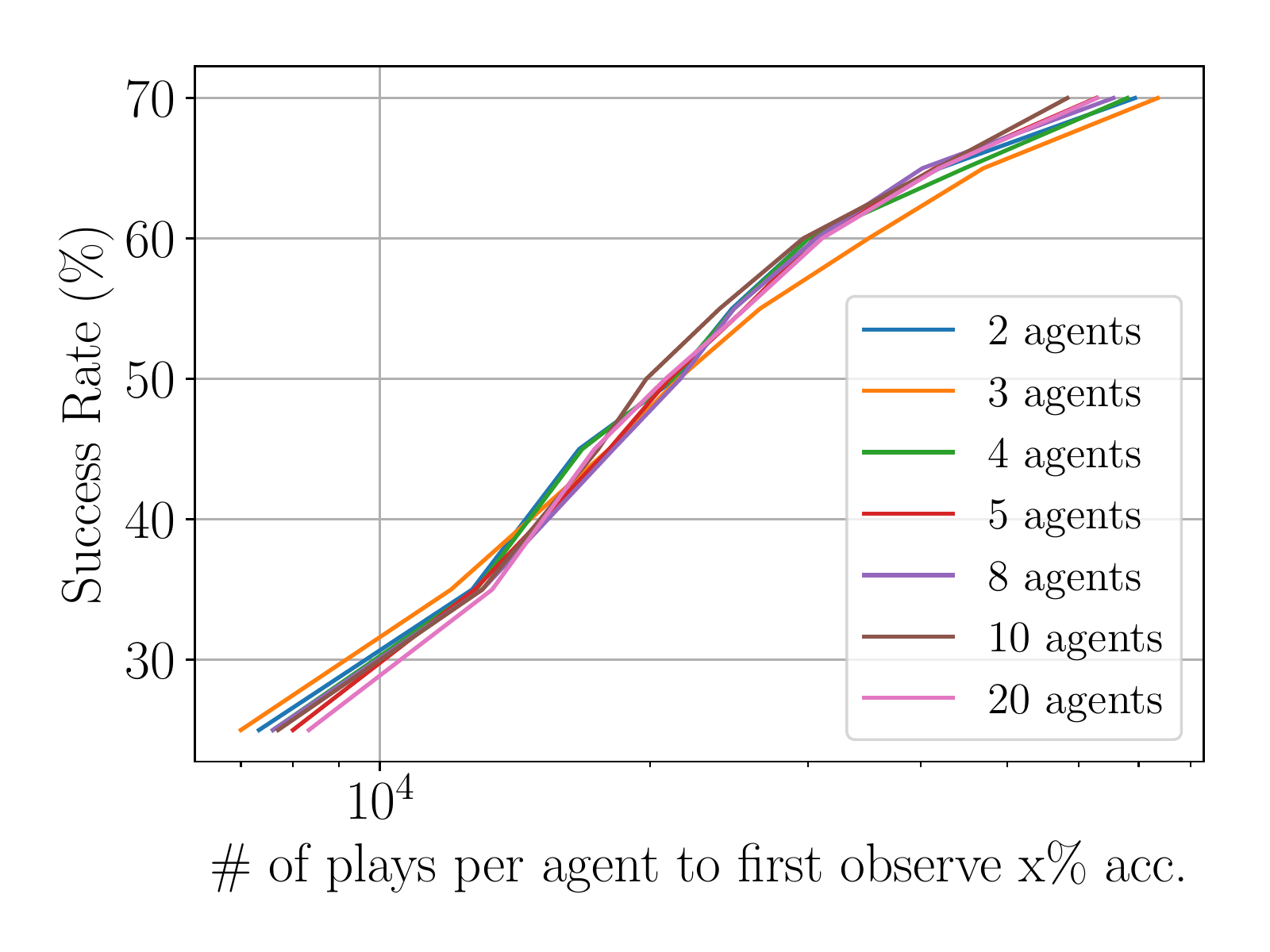}
(d)
\end{minipage}

\caption{
\label{fig:population}
Emergence of symmetric linguistic protocols (averaged over five runs). \textbf{(a)} Six communities were trained: we show the success rates under self-play and cross-play. We observe that at least three agents are necessary for the emergent protocol to be symmetric---without any specialized mechanism that enforces the symmetry of emergent protocol. 
\textbf{(b)} The average number of plays per agent to the first observed success rate $\geq70\%$ between a pair of agents in a linguistic community approximately stays constant with respect to the community size.
\textbf{(c)} The number of plays required for each agent in a linguistic community to reach the success rate of 75\% on average across all agents pairs in the community approximately stays constant with respect to the community size. 
\textbf{(d)} Each agent learns at approximately the same rate regardless of community size, which suggests that the emergent protocol emerges incrementally in a distributed manner rather than in a centralized way.
}
\end{figure*}

\paragraph{Training}

Each agent is trained using a hybrid of supervised and reinforcement learning. Each agent computes two predictive distributions before and after message exchange, $p^{\text{before}}(y|h)$ and $p^{\text{after}}(y|h)$, where $y$ refers to one of the $n_{\text{cap}}$ image captions and $h$ is the hidden state of the agent network, which fuses perceptual input with the last received message. Since we know the correct caption $y^*$ during training, we use supervised learning to train these two predictive distribution, $\max \log p^{\text{before}}(y^*|h) + \log p^{\text{after}}(y^*|h)$, using backpropagation and stochastic gradient descent~\cite{rumelhart1986learning}. Since messages are discrete, the message generating process cannot be backpropagated through. Instead, we use REINFORCE~\cite{williams1992simple} to maximize the reward $r$; $\max \mathbb{E}_{m|h,p(y|h)}\left[ r \right]$. 
We add the entropy of the message distribution as a regularization term, encouraging the exploration of various communication strategies in the early stage of learning.

\paragraph{Loss Functions}

There are four loss functions involved in each game. The first one is a prediction loss function. Given the index $y^*$ of a correct caption, the prediction loss function is
\begin{align*}
    \mathcal{L}_{\text{pred}} = -\log p(y=y^*|h).
\end{align*}
This loss is used twice based on the predictions before and after the message exchange; denoted as $\mathcal{L}_{\text{pred}}^{\text{before}}$ and $\mathcal{L}_{\text{pred}}^{\text{after}}$ respectively.

The second loss is a value loss function. After playing a game, the agent receives a reward $r$. The agent's value sub-network (see {\bf Fig.\ref{fig:agent}}) learns to predict this reward:
\begin{align*}
    \mathcal{L}_{\text{value}} = (r - V(h))^2.
\end{align*}

The third loss is a message loss function. During training, we sample one message $\tilde{m}$ from the message distribution generated by the agent $p(m|h,p(y|h))$. If this message led to a success, we increase the probability of the sampled message. Otherwise, we decrease it. The success is measured relative to the predicted value. The cost function is then:
\begin{align*}
    \mathcal{L}_{\text{msg}} = -(r-\hat{V}(h)) \log p(m=\tilde{m}|h,p(y|h)),
\end{align*}
where $\hat{V}(h)$ refers to using the predicted value but not updating the value sub-network according to this loss function. The gradient of this message loss function with respect to $p(m=\tilde{m}|h,p(y|h))$ corresponds to REINFORCE~\cite{williams1992simple}.

Lastly, we include an entropy penalty. Following \cite{evtimova2018emergent}, we encourage the entropy of the message distribution to be higher to facilitate exploration:
\begin{align*}
    \mathcal{L}_{\text{entropy}} = -\mathcal{H}(p(m|h,p(y|h))).
\end{align*}

\noindent The overall loss function is then the weighted sum of the four loss functions:
\begin{align*}
    \mathcal{L} = \alpha_{\text{pred}} \mathcal{L}_{\text{pred}} &+ \alpha_{\text{value}} \mathcal{L}_{\text{value}} \\
    &+ \alpha_{\text{msg}} \mathcal{L}_{\text{msg}} + \alpha_{\text{entropy}} \mathcal{L}_{\text{entropy}},
\end{align*}
where we set $\alpha_{\text{pred}}=1.0$, $\alpha_{\text{value}}=1.0$, $\alpha_{\text{msg}}=1.0$ and $\alpha_{\text{entropy}}=0.01$.

\section{Experiments}

\paragraph{When Are Protocols Symmetric?}

We first examine whether it is sufficient to have just two autonomous agents to develop a common communication protocol. That is, we ask whether a {\it symmetric} language emerges when there are only two agents in a linguistic community, or that they learn to speak distinct idiolects. We formally define ``mutual intelligibility'' in the communication protocol as the ability for each agent to play the game against itself. This is an elegant solution: if a shared communication protocol has emerged, the agent would not have any trouble playing a game with itself during test time (she ``understands'' what she ``says'' and ``says'' what she ``understands''). We run five simulations with random initialization and examine the success rate between two agents averaged over all the test examples, where the agents succeed on each example if both of them correctly guess the answer after communication.

\begin{figure*}[ht]
\begin{minipage}{0.32\textwidth}
\centering
\includegraphics[width=0.8\columnwidth]{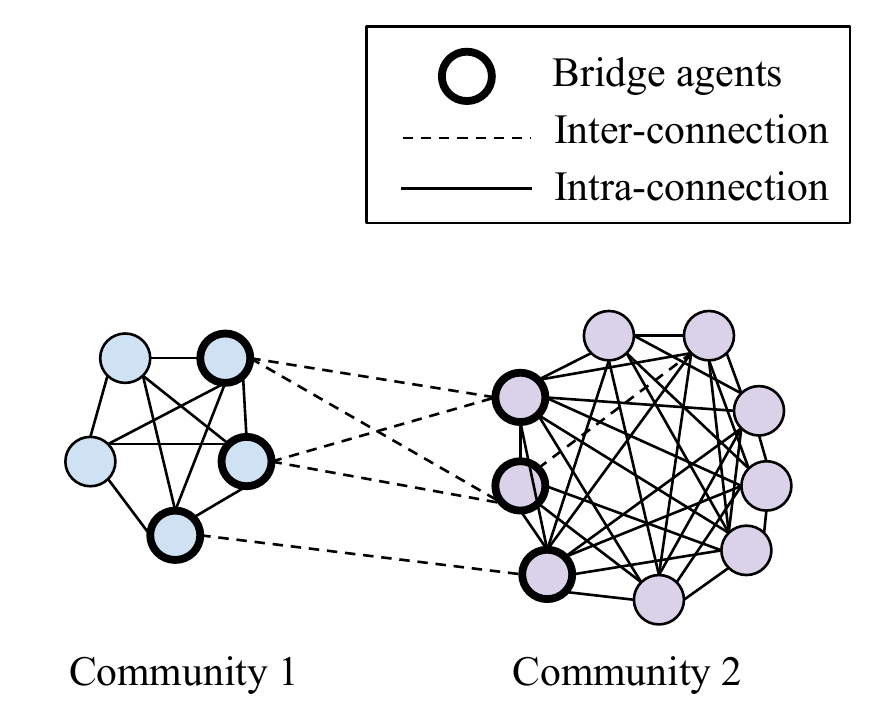}

(a)
\end{minipage}
\hfill
\begin{minipage}{0.32\textwidth}
\centering
\includegraphics[width=\columnwidth]{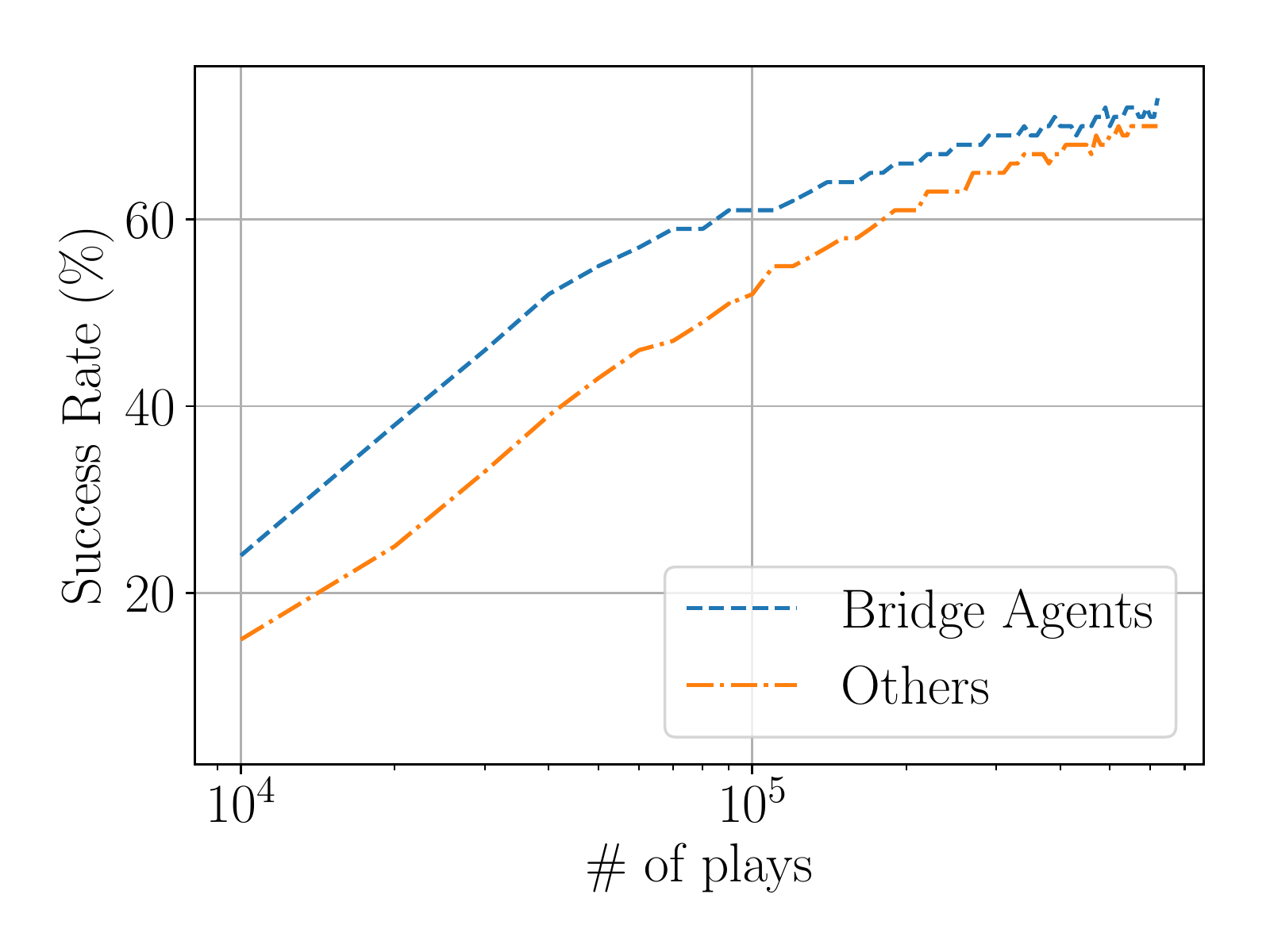}

(b)
\end{minipage}
\hfill
\begin{minipage}{0.32\textwidth}
\centering
\includegraphics[width=\columnwidth]{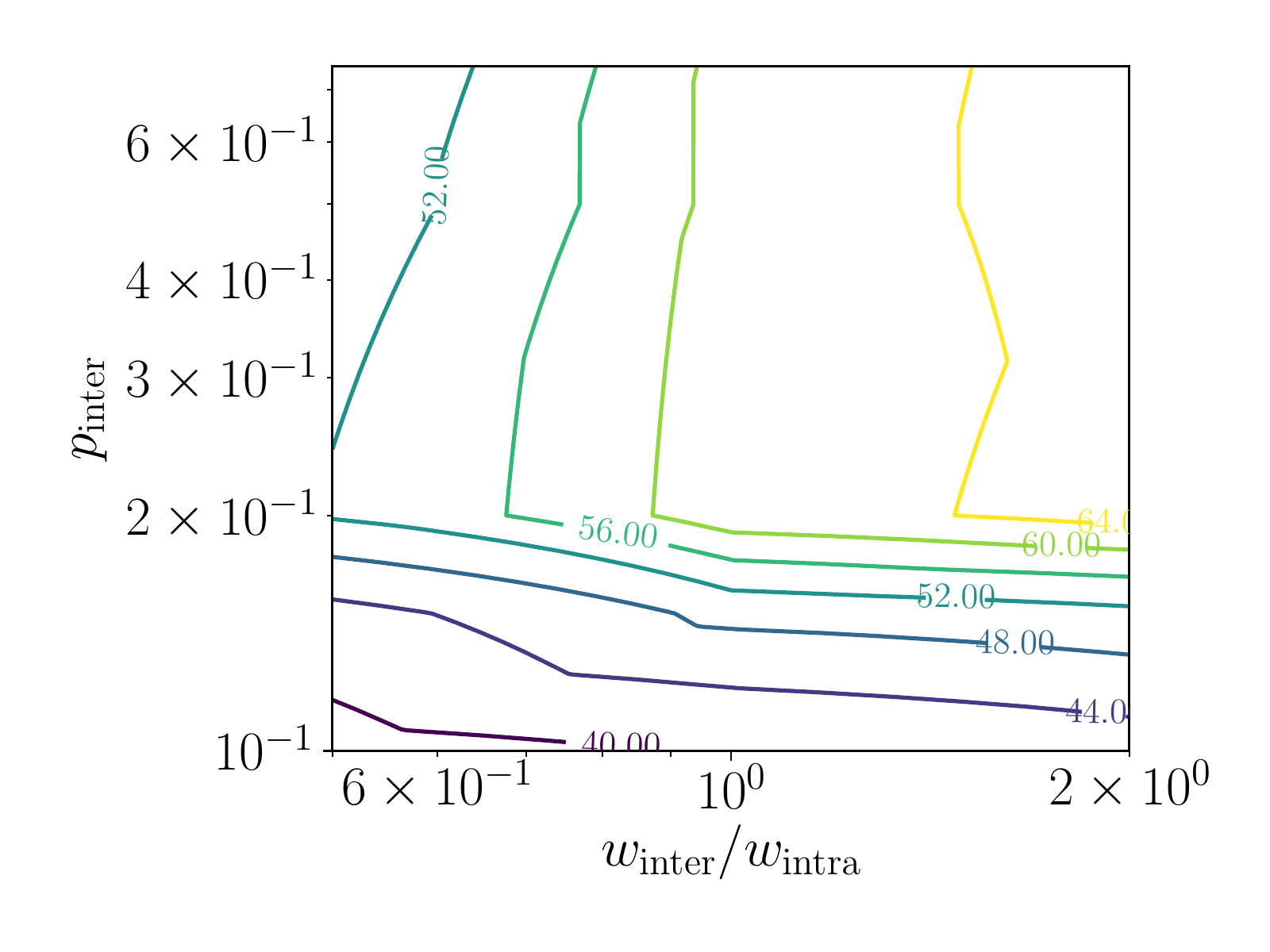}

(c)
\end{minipage}

\caption{\label{fig:bridge} 
The consequences of contact between two linguistic communities with initially distinct protocols. \textbf{(a)} Two linguistic communities come in contact. \textbf{(b)} Two communities of population ten each came in contact with the ratio of learning frequencies $(K ~w_{\text{inter}})/(L~ w_{\text{intra}})=1$ and the inter-group connectivity $p_{\text{inter}}=0.2$, after being separately trained in isolation (averaged  over  five  runs). The bridge agents, who interact with the agents from the other community, learn faster and better the new, shared emergent protocol. All the other agents however also rapidly learn to communicate with the agents from the other community, although they never interact directly with them.
\textbf{(c)} Contour plot visualization of the success rate after 200,000 plays after the contact by two linguistic communities while varying the ratio of learning frequencies $(K w_{\text{inter}})/(Lw_{\text{intra}})$ and the inter-group connectivity $p_{\text{inter}}$ (linearly interpolated from 15 experiments.) We observe that the success rate, which measures the level of convergence of two protocols, requires a certain level of the inter-group connectivity ($p_{\text{inter}} > 0.2$). Even when the inter-group connectivity is high enough, we further see that the bridge agents must interact with the agents from the other community enough ($(K w_{\text{inter}})/(Lw_{\text{intra}}) \ge 1$) for the converged protocol to be well understood by the agents from both communities. 
}
\end{figure*}

As shown in {\bf Fig.~\ref{fig:population}}~(a), two agents can play the game with a high success rate when they play with each other (cross-play). However, the success rate drops to random chance (10\%) when each agent plays against itself (self-play). This suggests that the emergent communication protocol is not symmetric: each agent has developed its own protocol, to which the other has adapted, leading to two incompatible idiolects, similar to previous observations in both fully~\cite{matignon2012independent} and partially observable settings~\cite{lanctot2017unified}.

We then run additional experiments having more than two agents, where every pair of agents $(v_i, v_j)$ interacts with an equal interaction intensity $w_{ij}=c$ (the success rate is averaged over all possible pairs of agents, unless stated otherwise). We notice that the success rates between self-play and cross-play are indistinguishable, strongly implying that a common, shared language emerges as a social convention if and only if we have more than two language users. This finding demonstrates that it is not strictly necessary to specifically equip an agent with an innate mechanism that ties listening and speaking, such as the obverter technique~\cite{oliphant1997learning,choi2018compositional}, nor any explicit community-wide coordination. All that is needed in order for a common language to emerge, at least within this framework, is a minimum number of agents.

We observe no detrimental effect to increasing the number of agents per linguistic community, even though more agents have to come to agree on a single protocol. As shown in {\bf Fig.~\ref{fig:population}}~(b--c), it takes approximately 60--65,000 plays per agent for us to observe the first instance of a pair of agents reaching the success rate of 70\% regardless of the community size. Similarly, it takes approximately 150-200,000 plays per agent for the success rate averaged over all pairs of agents in a community to reach 75\%, again, regardless of the size of the community. Surprisingly, the speed at which each agent learns stays constant with respect to the community size, as in {\bf Fig.~\ref{fig:population}}~(d). That is, we did not observe any correlation between community size and linguistic convergence of the entire community, which would probably only come into play with orders of magnitude more agents or for more complex reference games.

\paragraph{Understanding Convergence} 

\begin{figure*}[t]
\centering
\begin{minipage}{0.35\textwidth}
\centering
\includegraphics[width=\columnwidth]{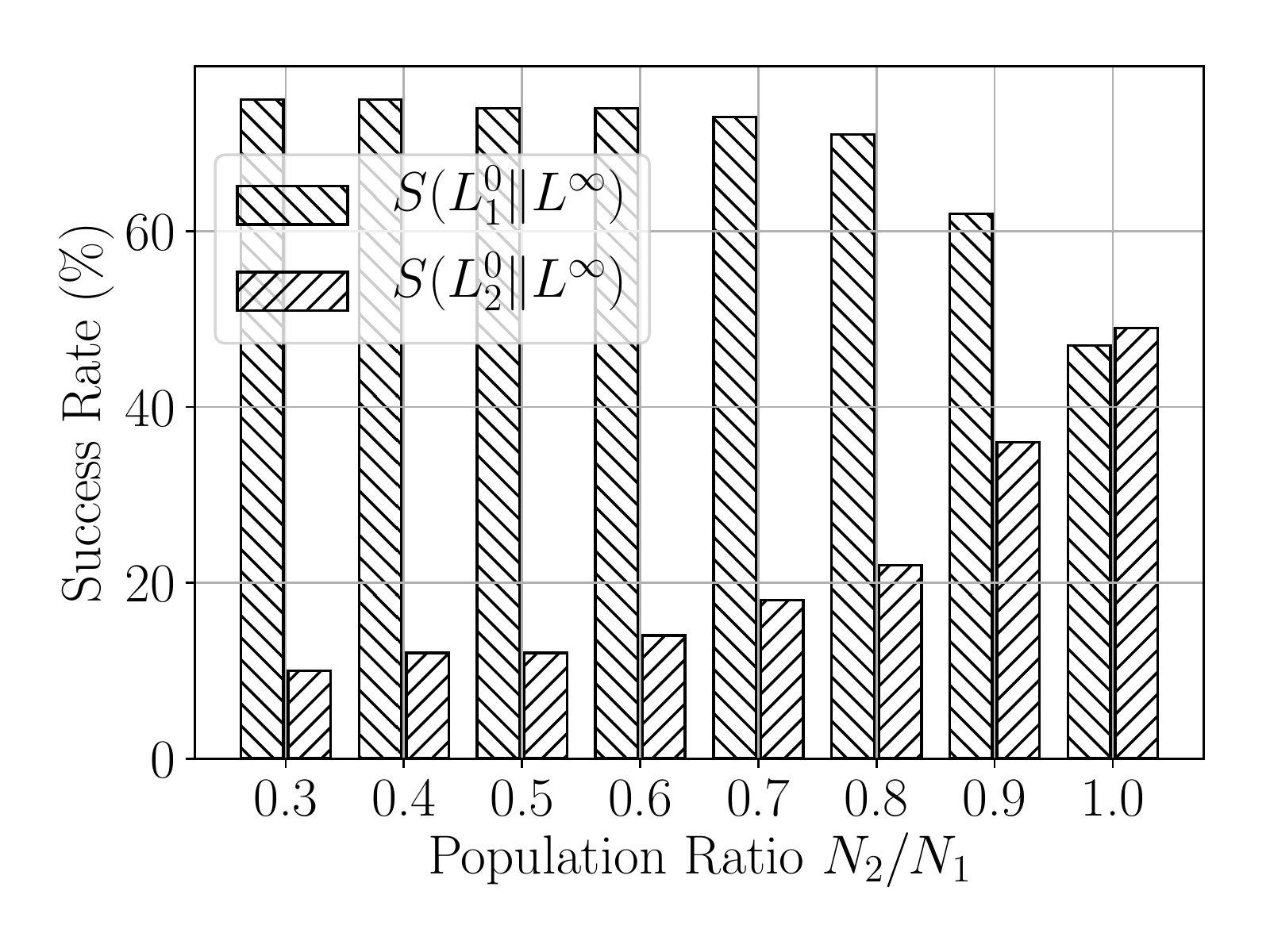}
(a)
\end{minipage}
\begin{minipage}{0.35\textwidth}
\centering
\includegraphics[width=\columnwidth]{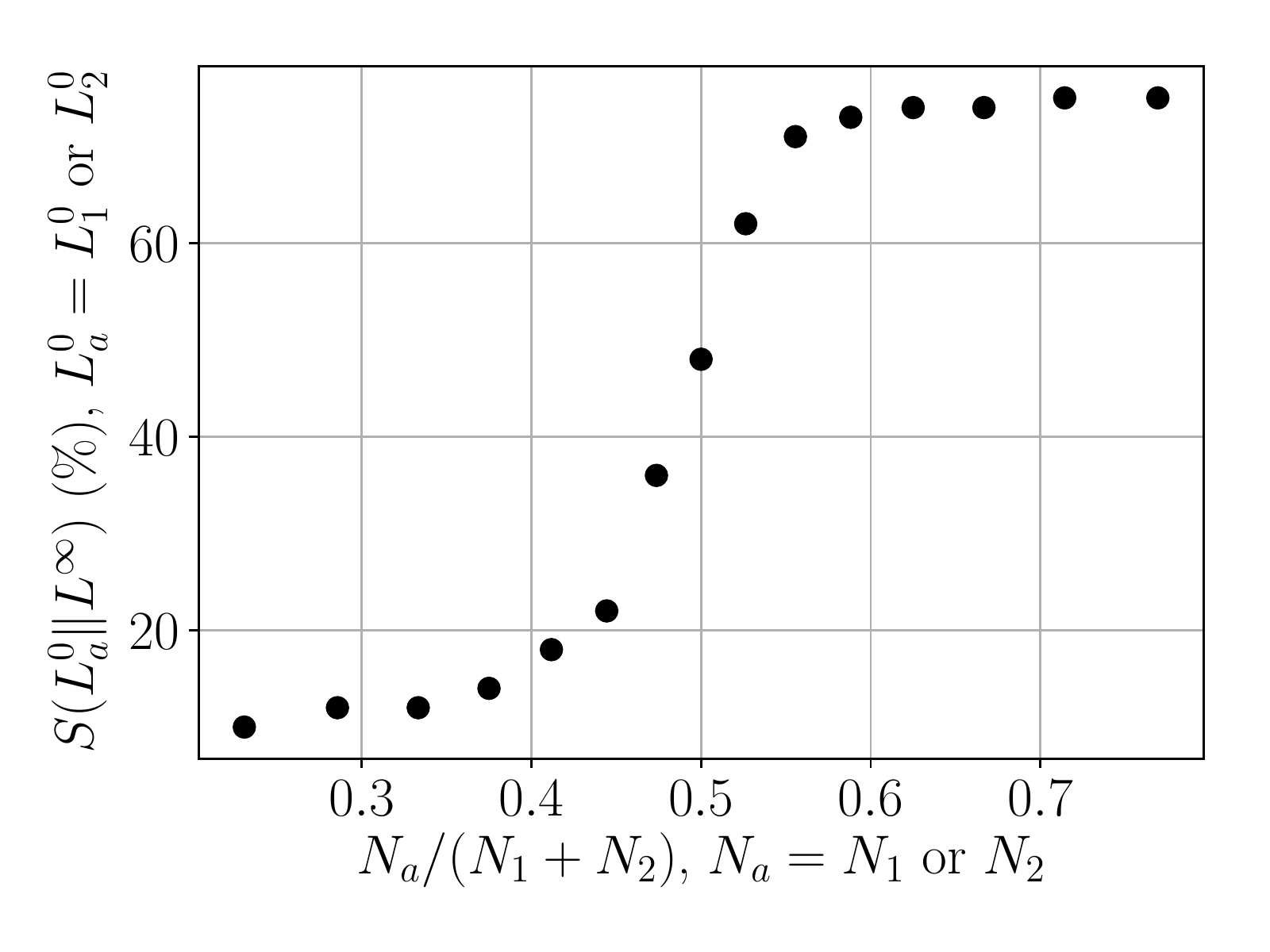}
(b)
\end{minipage}
\caption{\label{fig:birth} 
Relationship between the common emergent protocol and the original protocols after linguistic contact between two communities (averaged  over  five  runs). \textbf{(a)} Divergence of the common emergent protocol from the original protocols. Agents converge either to the majority protocol or to one in-between the two originals. \textbf{(b)} By varying the population ratio, it becomes clear that a near-even balance between two communities is necessary for a novel, contact protocol to emerge rather than the domination by a majority protocol.}
\end{figure*}

We next examine what happens when we expose different linguistic communities \emph{to each other}. Specifically, we consider two linguistic communities of population sizes $N_1$ and $N_2$, which are trained independently from each other as fully-connected communities and have developed separate communication protocols. We bring these two communities into contact with each other by introducing a new set of inter-community edges with probability $p_{\text{inter}}$ to form a new linguistic community. We assign a weight $w_{\text{inter}}$ to all the inter-group edges and another weight $w_{\text{intra}}$ to all the intra-group edges. We then examine how ``interaction intensity'' relates to language shift. See {\bf Fig.~\ref{fig:bridge}}~(a).

We first investigate two communities of identical population sizes ($N_1=N_2$) with the ratio of the learning frequencies of the intra-group pairs and inter-group pairs set to $(K w_{\text{inter}})/(L w_{\text{intra}})=1$, where $K$ is the number of inter-community connections and $L$ is the number of intra-community connections, and the inter-group connectivity chance set to $p_{\text{inter}}=0.2$.
We notice in {\bf Fig.~\ref{fig:bridge}}~(b) that the bridge agents learn to communicate better more rapidly (evident from the higher success rate among themselves), but the other agents quickly catch up (according to the success rate among themselves excluding the bridge agents), although these other agents never directly interact with agents from the other community. This finding demonstrates the rapid shift toward a common protocol in both groups where all agents learn to speak a shared language, regardless of whether they actually interact with agents from the other group.

Having established that linguistic contact leads to convergence of the communication protocol, we delve deeper into the impact of two major parameters, the ratio $(K w_{\text{inter}})/(L w_{\text{intra}})$ and the connectivity probability $p_{\text{inter}}$. We vary the ratio $(Kw_{\text{inter}})/(L w_{\text{intra}})$ of the learning frequencies of the intra-group pairs and inter-group pairs between $2/1$, $1/1$ and $1/2$, while fixing the inter-group connectivity to $p_{\text{inter}}=0.2$. After 200,000 plays, the former ($(K w_{\text{inter}})/(Lw_{\text{intra}})=2/1$) converges to a more tightly coupled linguistic community, achieving 65.6\% success rate between agents that never interacted with each other. On the other hand, when the inter-group interaction occurred only half as frequently as the intra-group interaction, the agents from the two groups can play together with a much lower 52.4\% success rate. We observed similar patterns over many different combinations of the ratio and inter-group connectivity. For example, we varied the inter-group connectivity $p_{\text{inter}}$ between $0.1$, $0.15$, $0.2$, $0.5$ and $0.75$ while the interaction ratio was fixed to $(K w_{\text{inter}})/(Lw_{\text{intra}})=2/1$. After 200,000 plays, we observed the success rates, averaged over all possible inter-group pairs, reach 42.1\%, 51.1\%, 65.55\%, 66.65\% and 66.3\%, respectively. This implies that there is a critical level of inter-group connectivity (around $0.2$ in this specific case) after which language propagation saturates.

In {\bf Fig.~\ref{fig:bridge}}~(c), we plot the interplay between the ratio $(Kw_{\text{inter}})/(Lw_{\text{intra}})$ and the inter-group connectivity $p_{\text{inter}}$ after interpolating from the fifteen experiments varying these parameters. This demonstrates that both parameters are important in determining the level of linguistic convergence.

\paragraph{Birth of a New Language: Emergence of a Contact Language}

We investigate the effect of the population size ratio $N_1/N_2$ between two linguistic communities when they come in contact. We study how population size is a factor in one language coming to ``dominate'' another language upon contact. We vary the ratio by fixing $N_1=10$ and varying $N_2 \in \left\{3,\ldots,10\right\}$. Each community is pretrained in isolation to develop its own protocol before being exposed to the other. We set the interaction ratio $(K w_{\text{inter}})/(Lw_{\text{intra}})$ to $1$ and the inter-group connectivity chance  $p_{\text{inter}}$ to $0.2$.

\begin{figure*}[ht]
\centering
\begin{minipage}{0.35\textwidth}
\centering
\includegraphics[width=\columnwidth]{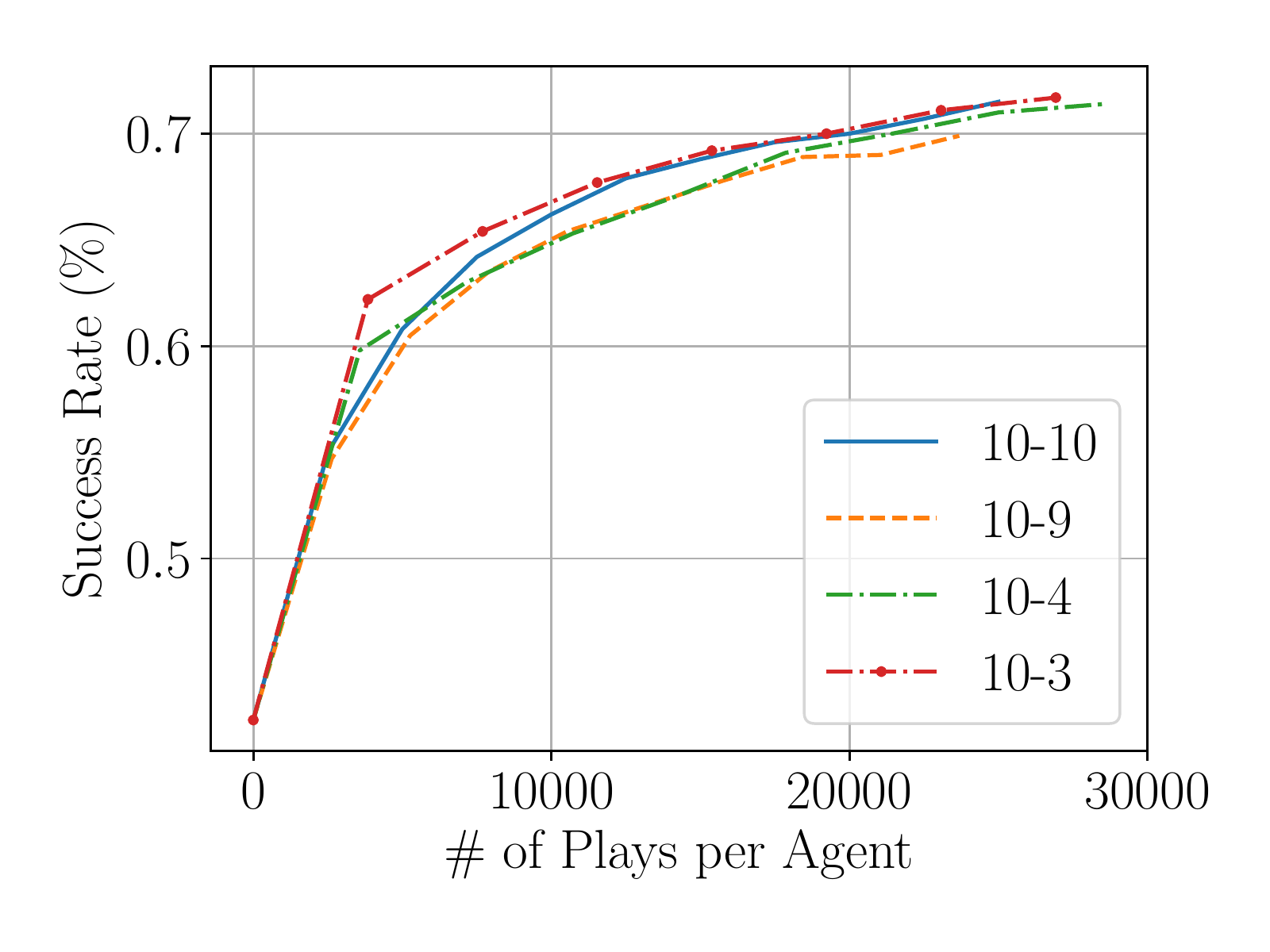}
(a) Success Rate
\end{minipage}
\begin{minipage}{0.35\textwidth}
\centering
\includegraphics[width=\columnwidth]{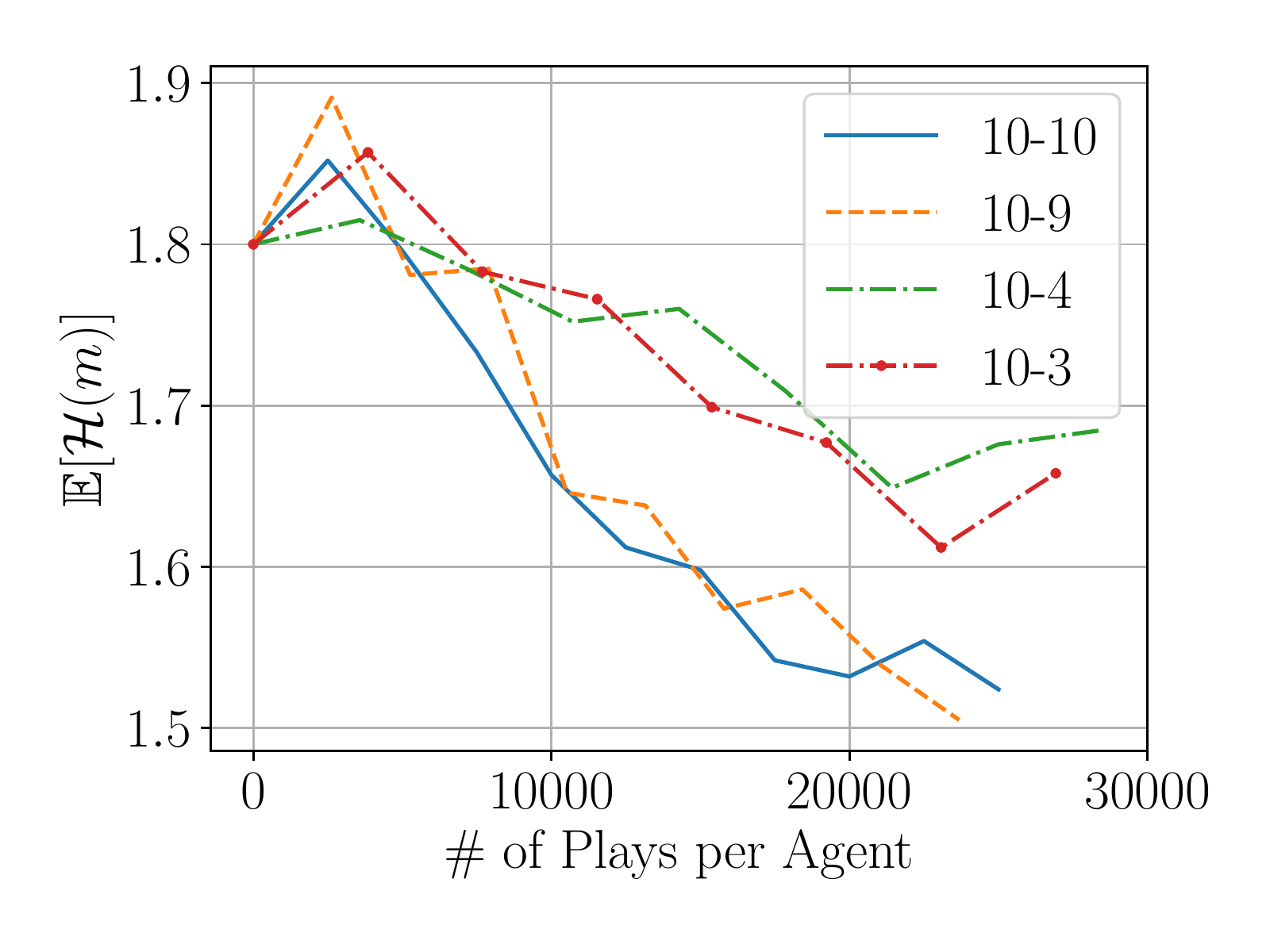}
(b) $\mathbb{E}[\mathcal{H}(p(m|h,p(y|h)))]$
\end{minipage}
\caption{\label{fig:complexity}
Evolution of success rate and protocol complexity as two communities of varying populations come into contact (averaged over three runs). Agents were finetuned until the average success rate reached at least $70\%$. The success rate evolves similarly in terms of the number of plays per agent. \textbf{(a)} We observe significantly different levels of complexity dependent on the population ratio. \textbf{(b)} The complexity is generally lower when the sizes of two communities are more balanced (10-10 and 10-9), while the complexity does not decrease as much when there is a significant imbalance in sizes between two communities (10-4 and 10-3).
}
\end{figure*}

We refer to the original protocols of the communities right after pretraining by $L_1^0$ and $L_2^0$. Each of these is then evolved further after these two communities come in contact, resulting in $L_1^{\infty}$ and $L_2^{\infty}$. The previous experiment on linguistic contact suggests that $L_1^{\infty} \approx L_2^{\infty}$ based on the fact that the agents from both communities can successfully play the game after coming in contact, so we refer to the final protocol as $L^{\infty}$. We examine how similar $L^{\infty}$ is to either of the original protocols, $L_1^0$ or $L_2^0$. This similarity is measured by letting the agent using $L_1^0$ or $L_2^0$ play against the one using $L^{\infty}$, which is naturally facilitated by the proposed framework. This ``historical self-play'' accuracy $S(L_{\cdot}^0\| L^{\infty})$ reflects the similarly of the original and final protocols. 

When the population ratio deviates from $N_1/N_2=1$, we observe that the final protocol rapidly converges to the majority protocol ($L_1^0$), evident from the near-perfect $S(L_1^0 \| L^{\infty})$ and the near-chance $S(L_2^0 \| L^{\infty})$ in {\bf Fig.~\ref{fig:birth}}. This results from the fact that members of both communities are rewarded for cooperating and playing the game well (via the bridge agents). In other words, the agents prefer to integrate or assimilate rather than segregate, similar to how it has been found that minority groups shift toward the use of dominant language~\cite{fase1992maintenance}. On the other hand, we observe $S(L_1^0 \| L^{\infty}) \approx S(L_2^1 \| L^{\infty})$ and that both of these historical self-play accuracies are significantly above chance, when the population ratio is closer to or exactly $1$. It is impossible to identify either $L_1^0$ or $L_2^0$ as an ancestor of $L^{\infty}$, but $L^{\infty}$ is rather a combination of these two original protocols, which is a key characteristic of contact languages \cite{matras2009language}. Both of these observations suggest the potential of the proposed framework for simulating and understanding the birth and death of new languages via linguistic contact.


We further investigate the complexity of the contact language arising from two linguistic communities coming into contact. We define complexity as the uncertainty of an agent when generating a message, and measure the entropy of the message distribution $\mathcal{H}(p(m|h,p(y|h)))$. Higher entropy indicates that agents can express states in many different ways: in other words, the more complex a language, the higher the degree of freedom. For each linguistic community, we then compute the average of their message distributions in order to characterize the complexity of a learned communication protocol.

We observe in {\bf Fig.~\ref{fig:complexity}}~(b) that the complexity decreases when two communities come into contact. This observation agrees with a similar phenomenon of structural simplification in creole languages which are understood to arise from the contact of two or more languages~\cite{parkvall2008simplicity,bakker2011}.
We also observe that the complexity plateaus earlier when there is a larger imbalance between two communities' population (10-3 and 10-4), while it drops further with more balanced communities (10-9 and 10-10). This implies that the new-born contact languages arising from the contact of two similarly-sized communities tend to be substantially simpler.

\begin{figure*}[ht]
\begin{minipage}{0.19\textwidth}
\centering
\includegraphics[width=\columnwidth]{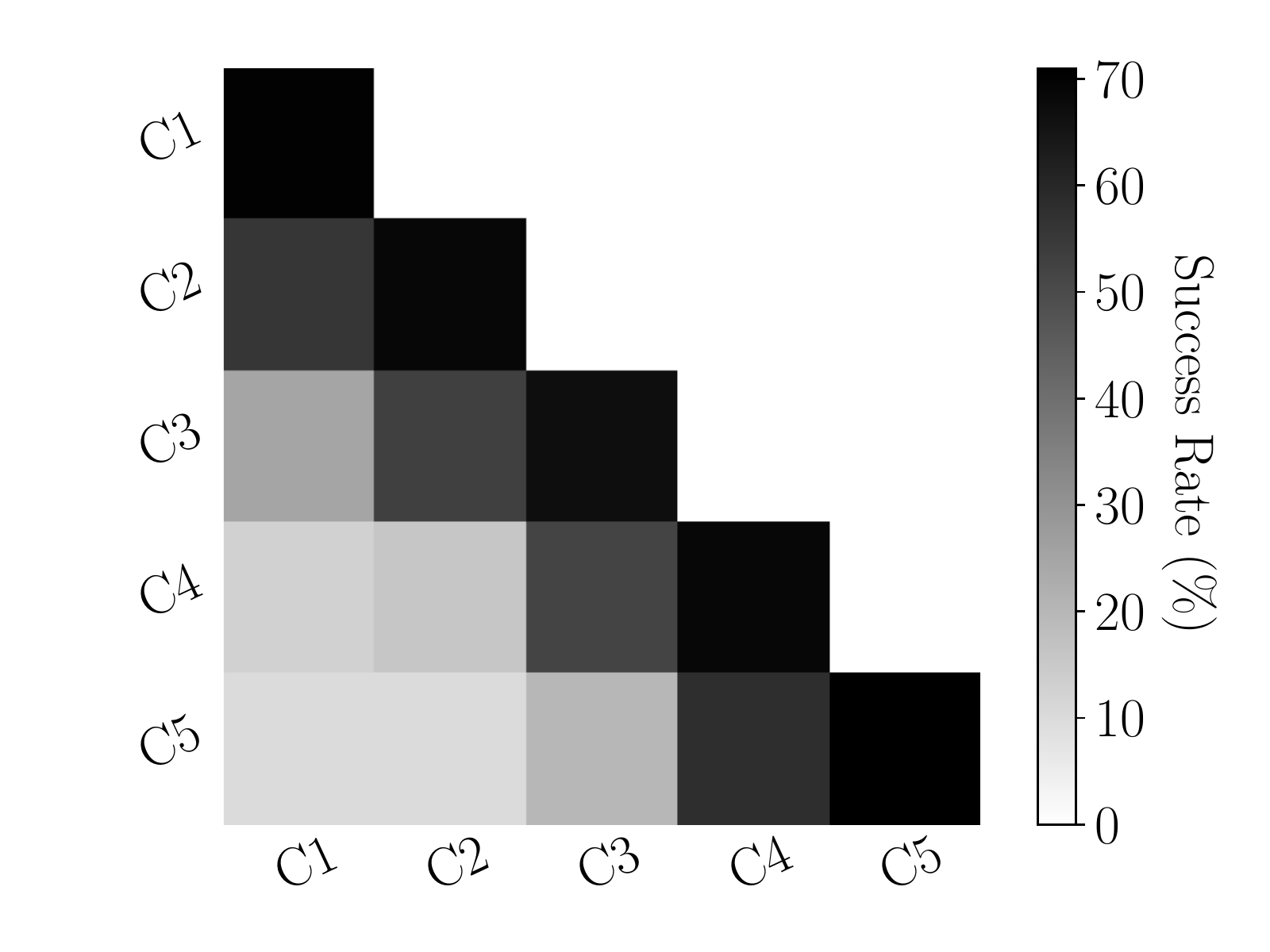}
(a) 5-5-5-5-5
\end{minipage}
\begin{minipage}{0.19\textwidth}
\centering
\includegraphics[width=\columnwidth]{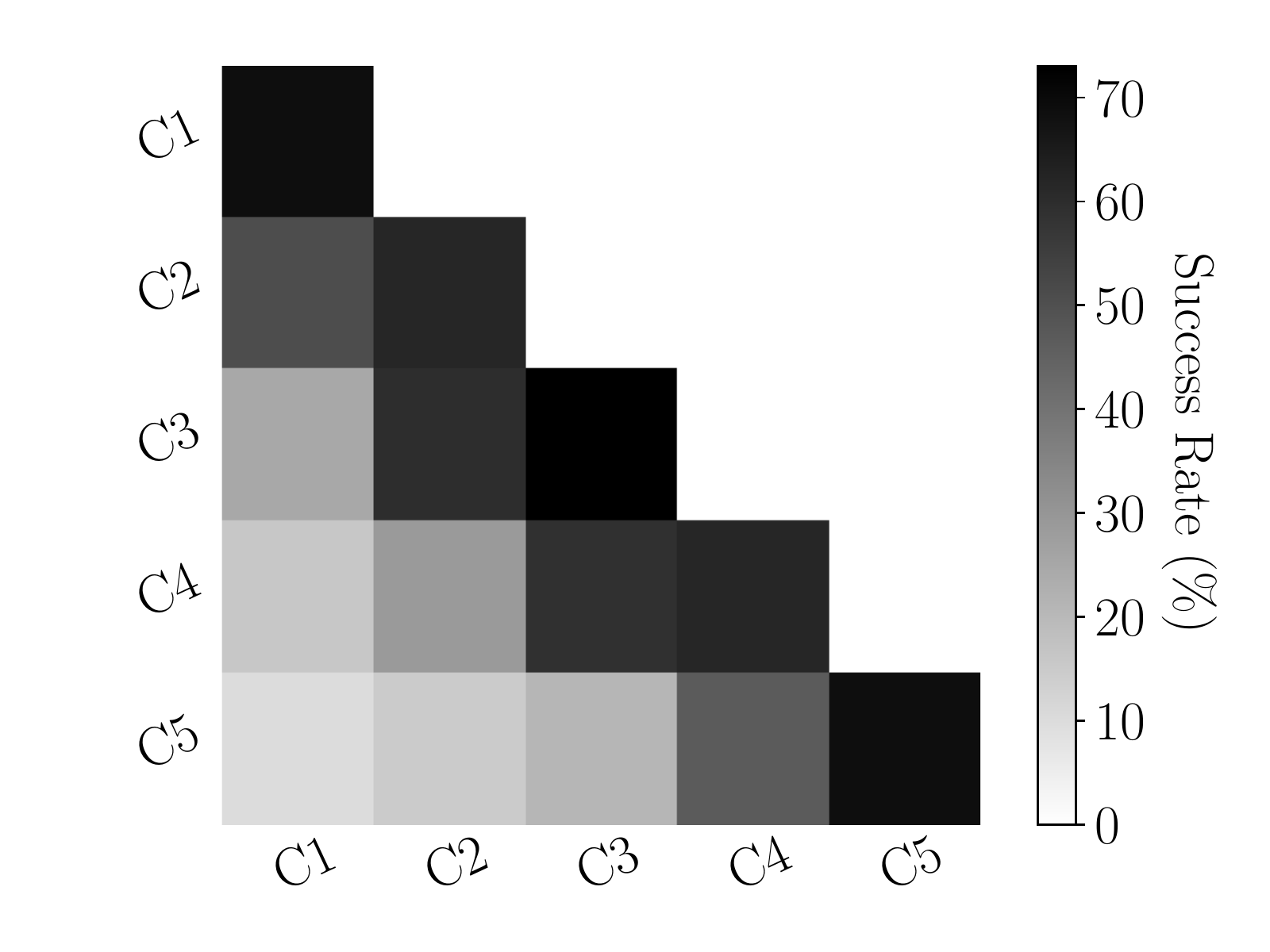}

(b) 5-5-10-5-5
\end{minipage}
\begin{minipage}{0.19\textwidth}
\centering
\includegraphics[width=\columnwidth]{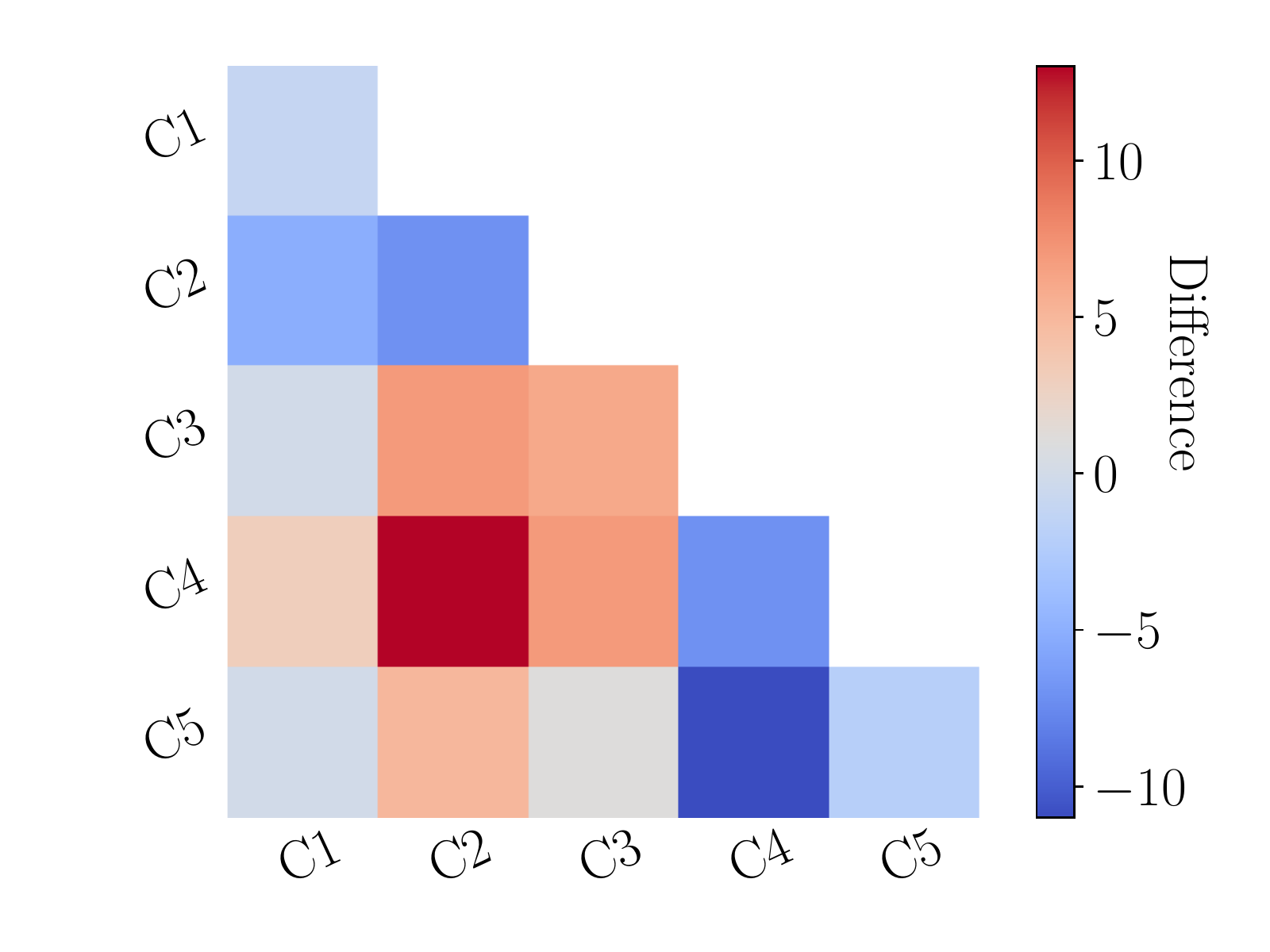}

(c) Difference (b)-(a)
\end{minipage}
\begin{minipage}{0.19\textwidth}
\centering
\includegraphics[width=\columnwidth]{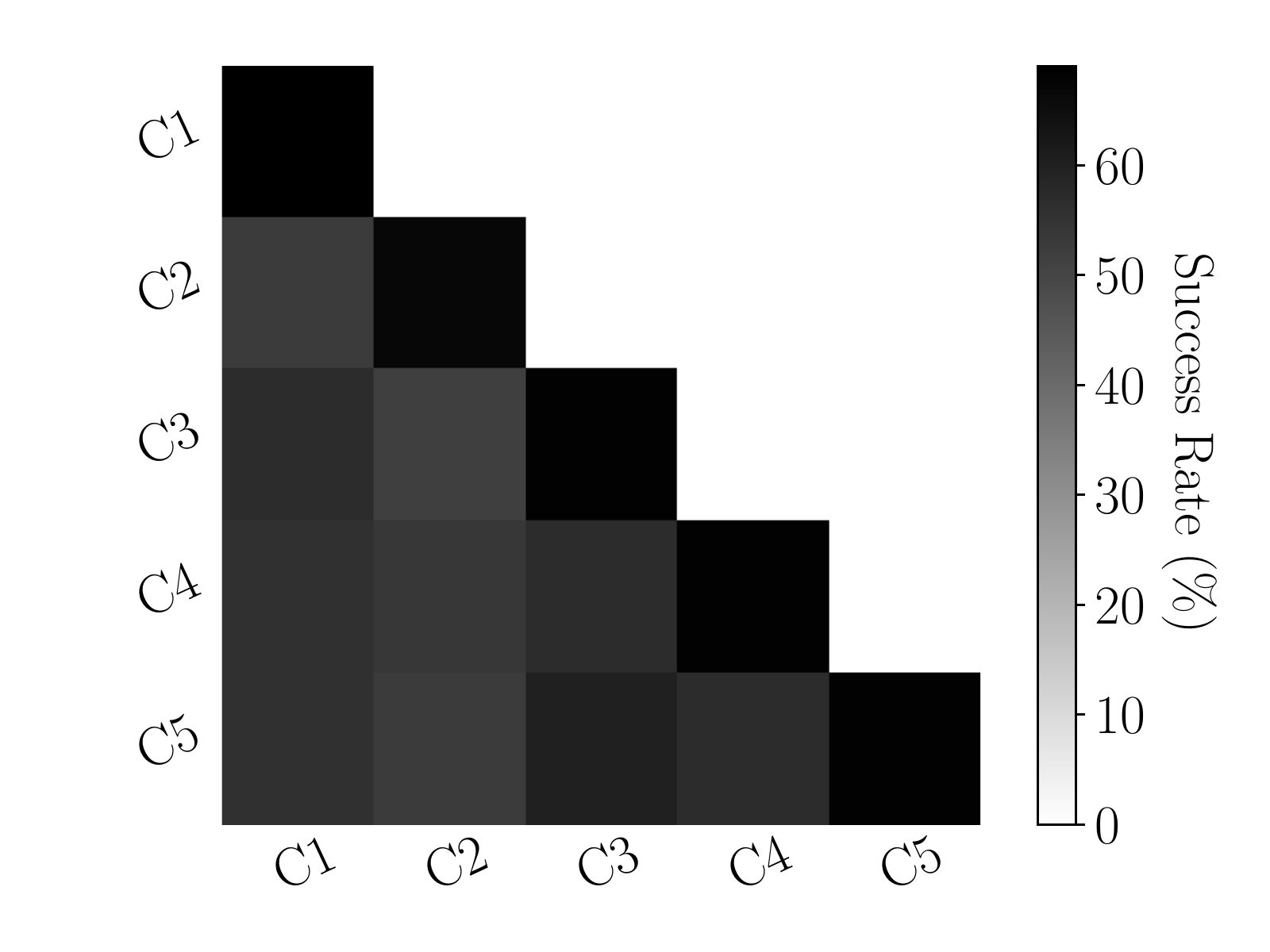}

(d) Dense (a)
\end{minipage}
\begin{minipage}{0.19\textwidth}
\centering
\includegraphics[width=\columnwidth]{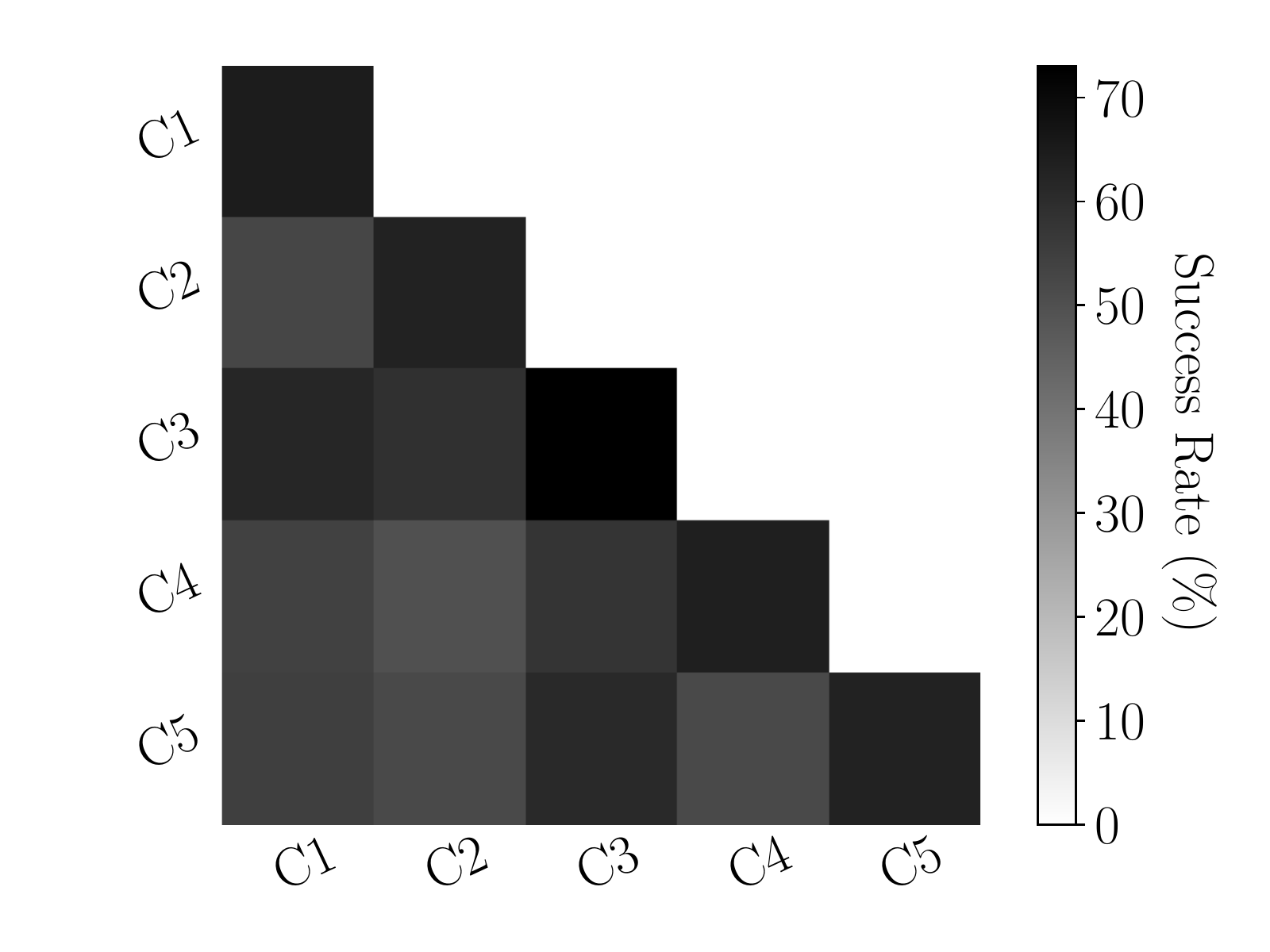}

(e) Dense (b)
\end{minipage}

\caption{\label{fig:chain} 
We plot the protocol similarities among five communities in a chain (averaged  over  three  runs). \textbf{(a)} Neighbouring communities exhibit higher protocol similarities, but distant communities are not mutually intelligible. \textbf{(b)} With a larger community in the middle of a chain, we observe higher levels of protocol similarities among the communities in the chain. \textbf{(c)} Differences between the two, showing that the protocols near the center become more similar to each other when the center community dominates, at the expense of intelligibility between further-removed communities. \textbf{(d, e)} We do not observe a continuum when the communities are connected densely.
}
\vspace{-3mm}
\end{figure*}

\paragraph{A Linguistic Continuum of Contact}

We generalize the previous setting to having $M>2$ linguistic communities in various topologies. We start by pretraining $M$ linguistic communities of populations $N_1,N_2, \ldots, N_M$ respectively, evolving $M$ distinct communication protocols. We then chain them such that each consecutive pair, $C_i$ and $C_{i+1}$, comes in contact with a pre-specified inter-group connectivity chance $p_{\text{inter}}$ and interaction ratio $(Kw_{\text{inter}})/(Lw_{\text{intra}})$, and begin training all of the communities jointly. We study the emergence of a linguistic continuum, similar to the dialect continuums that can be found in natural languages such as the Nordic Germanic dialects of Scandinavia ~\cite{chrystal1987cambridge}. Often, speakers on the border are mutually intelligible, while those from communities geographically separated by many intermediate ones cannot communicate.

We start by considering a chain of five communities of equal population ($M=5$). As plotted in {\bf Fig.~\ref{fig:chain}}~(a), we clearly observe the emergence of a linguistic continuum. The agents from a pair of adjacent communities can communicate with each other almost as well as those within a single community, while communicability rapidly degrades as the distance between a pair of communities grows (off-diagonal). The agents from $C_1$ and $C_5$ cannot understand each other at all, achieving the near-chance success rate. A similar continuum is observed when we increased the population of the center community two-fold (5$\to$10). This continuum however exhibits properties different from the original chain of equal-sized communities: the center communities $C_2$, $C_3$ and $C_4$ become more tightly coupled, as evident from the higher success rate among those in {\bf Fig.~\ref{fig:chain}} (b-c). This however happens at the cost of communicability between the agents from furthest-removed communities.

To see if the emergence of such a continuum is due to topological properties of communities, we show similarities $S(L_i^{\infty} \| L_j^{\infty})$ among the five communities when densely connected in {\bf Fig.~\ref{fig:chain}} (d-e). Unlike for chaining, we ensure that every pair of communities comes in contact with each other in a densely connected topology. All communities are uniformly similar, confirming that the linguistic continuum arises from the topology.

\section{Discussion and Conclusion}

We have described a framework for the large-scale investigation of complex linguistic phenomena via multi-agent communication games.
We started by observing that a symmetric communication protocol emerges without any innate, explicit mechanism built in an agent, when there were three or more of them in a community. We then demonstrated the emergence of several complex linguistic phenomena in this simple framework.

First, the result of linguistic contact between communities is determined by inter- and intra-group connectivity patterns. Given sufficient inter-group connectivity, languages become mutually intelligible through contact, even for agents that have not been exposed to the other language. Second, linguistic contact over time either converges to the majority protocol, leading to the extinction of the other language, or gives rise to an original ``creole'' protocol that has lower complexity than the original languages, if the communities are balanced. Third, a linguistic continuum emerges, where neighboring languages are more mutually intelligible than farther removed languages. The topology of the continuum governs its behavior, and a very dominant central language causes its neighbors to lose mutual intelligibility with communities that are not directly exposed to that central language.

We conclude that intricate properties of language evolution need not depend on complex evolved linguistic capabilities, but can emerge from simple social exchanges between perceptually-enabled agents playing communication games. Language evolution and its properties can be effectively and efficiently investigated in-depth under the proposed framework.

In future work, it would be useful to investigate more complicated environments and more complex agent interactions. The setting in this paper included a very simple vision task, and we suspect emergent linguistic phenomena would be more pronounced and even more interesting to study in more sophisticated settings.

\section*{Acknowledgments}

We thank the anonymous reviewers for providing helpful feedback. We thank Marco Baroni, Alex Peysakhovich and Adina Williams for their comments on an earlier draft of this work.

\bibliography{emnlp-ijcnlp-2019}
\bibliographystyle{acl_natbib}


\end{document}


\maketitle

\section{Communication Game}

The game is similar to other games in the language evolution literature \cite{nowak1999evolution,nowak1999evolutionary}, such as the naming game~\cite{steels1995self,baronchelli2008depth}, the guessing game~\cite{steels2015talking} and the category game \cite{puglisi2008cultural}. Unlike the guessing and category games, the proposed game is symmetric between the participating agents and is partially observed. Unlike the naming game, the agents in the proposed game can handle sensory input and learn to capture sophisticated relationships between objects with arbitrary messages by means of supervised and reinforcement learning.

\section{Agent}

\subsection{Design}
A reference agent is implemented as a deep neural network consisting of multiple component sub-networks, based on recent advances in deep learning~\cite{lecun2015deep}.

Each agent is equipped with visual perception and the ability to communicate, both of which are implemented jointly in a single deep neural network and trained end-to-end using reinforcement learning to play the proposed communication game. The sensory sub-network is implemented as a ResNet-34, the state-of-the-art deep convolutional network from \cite{he2016deep}, with fixed weights (i.e., using the weights obtained from training on ImageNet classification). We transfer the final pre-classification layer in order to extract a 512-dimensional feature vector from the partially-visible input image, which is further transformed with a trainable dense layer and ReLU~\cite{nair2010rectified,glorot2011deep} activation function to a 100-dimensional feature vector, $h_{\text{sensor}}$. The receiver sub-network is a recurrent neural network based on gated recurrent units~\cite{cho2014learning} and is able to process multiple turns of message exchanges. It encodes the history of received binary-vector messages into a 100-dimensional feature vector $h_{\text{message}}$. These two vectors are then combined using the fusion sub-network into a single 100-dimensional vector $h$ which represents the agent's {\it internal state}. Based on this internal state $h$, the agent computes three quantities. First, the predictor sub-network computes the predictive distribution $p(y|h)$ over all the captions by comparing the internal state $h$ against the feature vector $h_{\text{cap}}^i$ of each caption outputted by the text sub-network. The most likely caption under this distribution is the agent's answer. Second, the sender sub-network computes the distribution $p(m|h, p(y|h))$ of a message to be sent, using the output of the predictor sub-network to incorporate the agent's current view of which caption is correct. During training, the agent stochastically samples a binary-vector message from this distribution $\tilde{m} \sim m|h,p(y|h)$, and during test, it uses the most likely message $\hat{m} = \arg\max_m \log p(m|h, p(y|h))$. The text sub-network encodes predictions (e.g. ``there is a red circle'') as representations of natural language. Lastly, the reward is predicted by the value sub-network, which is only used during training to 
stabilize learning.

This modular agent design allows us to easily swap various sub-networks with other architectures within the same framework. For instance, by replacing the sender sub-network with a recurrent neural network, the agent can generate a sequence of symbols rather than a binary vector. One could also modify the game to include other sensory modalities by modifying the sensory sub-network.


\subsection{Detailed specification}

The agent extracts a sensory feature vector $h_{\text{img}} \in \mathbb{R}^{512}$ using a pretrained ResNet-34 excluding the final classification layer, denoted ResNet-34$^-$, which is available from torchvision package\footnote{
\url{https://pytorch.org/docs/stable/torchvision/index.html}
}
from PyTorch.\footnote{
\url{https://pytorch.org/}
}
It is followed by
\begin{align*}
    h_{\text{img}} = \text{ReLU}(W_{\text{img}} \text{ResNet-34$^-$}(x) + b_{\text{img}}),
\end{align*}
where $\text{ReLU}(a) = \max(0, a)$ is a rectified linear unit, and $W_{\text{img}} \in \mathbb{R}^{100 \times 512}$ and $b_{\text{img}} \in \mathbb{R}^{100}$ are trainable parameters. 

A message $m \in \left\{ 0, 1\right\}^8$ is processed by a gated recurrent unit (GRU, \cite{cho2014learning}) each time a new message is received:
\begin{align*}
    h_{\text{msg}} \leftarrow \text{GRU}(m, h_{\text{msg}}) \in \mathbb{R}^{100},
\end{align*}
where the GRU's hidden state, $h_{\text{msg}}$, is initialized to zero at the beginning of each game.
In this manuscript's setup, there is only one message received per agent per game. The game begins with each agent receiving a blank message (all zeros) and making a prediction about the correct caption. Next the agent selected to communicate first
sends a message to the other agent, who after receiving it sends a message back to the first agent.\footnote{
The order of message exchange is randomized each play.
} 
Finally, each agent again tries to predict the correct caption.

The image and message vectors, $h_{\text{img}}$ and $h_{\text{msg}}$, are concatenated and combined into a single vector by the fusion sub-network by
\begin{align*}
    h = U_{\text{fuse}} [h_{\text{img}}; h_{\text{msg}}] + b_{\text{fuse}},
\end{align*}
where $U_{\text{fuse}} \in \mathbb{R}^{100 \times 200}$ and $b_{\text{fuse}} \in \mathbb{R}^{100}$ are trainable parameters. This fused vector $h$ is used to represent the agent's internal state. 

Each candidate caption $c$ is turned into a vector by the text sub-network:
\begin{align*}
    h_{\text{desc}}^{c} = \frac{1}{T_c}\sum_{t=1}^{T_c} E[w_t^c],
\end{align*}
where $w_t^c$ is the $t$-th word of the candidate caption $c$, $E: V \to \mathbb{R}^{100}$ is a trainable word embedding function with the vocabulary $V$, and $T_c$ is the length of the caption. We build the embedding function $E$ using a set of pretrained 100-dimensional GloVe~\cite{pennington2014glove} vectors. 
The predictor sub-network then compares each candidate caption against the fused vector to compute its score:
\begin{align*}
    \alpha_c = h^\top h_{\text{desc}}^{c},
\end{align*}
These scores are normalized to become a probability~\cite{bridle1990training}:
\begin{align*}
    p(y=c|h) = \frac{\exp(\alpha_c)}{\sum_{c'=1}^{10} \exp(\alpha_{c'})}.
\end{align*}

Given the fused vector, the agent computes the message to be sent to the partner. This is done by first computing the message distribution, using the normalized probabilities from the predictor sub-network to incorporate the agent's current belief about the correct caption. Assuming a $L=8$-dimensional binary message as done in this paper, the distribution is computed by first calculating a weighted sum of the caption;
\begin{align*}
    h_{\text{desc}}^{w} = \sum_{i=1}^{10} \alpha_c~ h_{\text{desc}}^{c}
\end{align*}
This is combined with the hidden state to generate the message distribution;
\begin{align*}
    & h_m = \tanh(h^\top w^l_{\text{gen\_h}} + {h_{\text{desc}}^{w}}^\top w^l_{\text{gen\_d}} + b^l_{\text{gen\_h}} + b^l_{\text{gen\_d}}),\\
    & p(m_l=1|h) = 1/(1+\exp(-h_m^\top w^l_{\text{gen\_m}} - b^l_{\text{gen\_m}})) \in \left(0, 1\right),
\end{align*}
where $w^l_{\text{gen\_h}}$, $b^l_{\text{gen\_h}}$, $w^l_{\text{gen\_d}}$,  $b^l_{\text{gen\_d}}$, $w^l_{\text{gen\_m}}$ and $b^l_{\text{gen\_m}}$ are trainable parameters. During training, we sample from this distribution, while we simply round the probability for each bit at test time.

In order to reduce the variance of policy gradients, we use a learned value estimate as a baseline. The agent estimates the expected reward/return given the observation--image and message--using the value sub-network. It takes as input the fused vector and outputs a single scalar:
\begin{align*}
    V(h) = w_{\text{v\_2}}^\top ~ \max(0, w_{\text{v\_1}}^\top h + b_{\text{v\_1}}) + b_{\text{v\_2}} ,
\end{align*}
where $w_{\text{v\_1}}$, $b_{\text{v\_1}}$, $w_{\text{v\_2}}$ and $b_{\text{v\_2}}$ are trainable parameters. 

\section{Data Generation}

We modify ShapeWorld~\cite{kuhnle2017shapeworld} to generate a set of training, validation and test examples. Each example is a 128$\times$128 RGB image containing an object with a simple shape and color. There are eight shapes--`circle', `cross', `ellipse', `pentagon', `rectangle', `semicircle', `square' and `triangle'-- and seven colors--`blue', `cyan', `gray', `green', `magenta', `red' and `yellow'. The size and position of the object in an image are randomly decided, while ensuring that the object size is relatively small compared to the image size. Each image is associated with a textual caption, i.e., a sentences which describes the shape, color, or shape and color of the object. Some examples include ``there is a blue square'', ``there is a yellow shape'' and ``there is a square''. 

We randomly select nine captions from the other images in order to create 10 candidate captions from which the correct one must be selected by both agents.
$\sim$13-16\% of examples are ambiguous due to the fact that objects can be described by their shape, color, or shape and color.

Each image is partitioned into two parts, each of which is shown to only one of two players. Due to the small size of the object in each image and random partitioning, the object is only visible to one of the players in approximately 82-84\% of images. When the object is split into two partitions, both agents may be able to correctly solve the problem without consulting the other (yet a split rectangle may be wrongly perceived as a triangle, for instance). It is necessary for the agents to communicate in all the other cases. In {\bf Fig.~1}~(a), we show example image partitions. Random partitioning happens during training and at evaluation time without any fixed partition per image.

We create 5,000 training examples while excluding the following combinations: `red square', `green triangle', `blue circle', `yellow rectangle', `magenta cross' and `cyan ellipse'. These were excluded in order for us to test the generalization of trained agents to unseen combinations of color and shape.\footnote{
    This is done to facilitate future research, and we do not test this generalization property in this paper.
}
We similarly construct 1,000 in-domain evaluation examples with only combinations included in the training set, and 5,000 out-of-domain evaluation examples which contain all possible combinations. Both of these are held-out during training and are used for evaluation, with the in-domain results reported throughout this paper. Out of these 6,000 examples, 271 combinations of shape and color do not appear in the training set.

\section{Learning}

\subsection{Optimization}

We use stochastic gradient descent with minibatch size of 32 and use RMSProp to automatically adapt per-parameter learning rates.

\section{Code and Data}


The code used for implementing the proposed framework as well as the experiments in this manuscript is publicly available at \url{https://github.com/lgraesser/MultimodalGame}. The generated data used in the experiments can be downloaded from
\url{https://goo.gl/HgHV1H}.

\bibliography{emnlp-ijcnlp-2019}
\bibliographystyle{acl_natbib}
